\pdfoutput=1

\documentclass[11pt]{article}

\usepackage{ACL2023}

\usepackage{times}
\usepackage{latexsym}

\usepackage[T1]{fontenc}

\usepackage[utf8]{inputenc}

\usepackage{microtype}

\usepackage{inconsolata}
\usepackage{hyperref}       %
\usepackage{url}            %
\usepackage{booktabs}       %
\usepackage{amsfonts}       %
\usepackage{nicefrac}       %
\usepackage{microtype}      %
\usepackage{xcolor}         %

\usepackage{times}
\usepackage{latexsym}
\usepackage[utf8]{inputenc}
\usepackage{inconsolata}
\usepackage{microtype}
\usepackage{graphicx}
\usepackage{booktabs} %
\usepackage{enumitem}
\usepackage{tikz}
\usepackage{pgfplots}
\usepackage{tabulary}
\usepackage{tabularx}
\usepackage{array}
\usepackage[noend]{algpseudocode}
\usepackage{bbm}
\usepackage{indentfirst}
\newcommand{\abbr}[0]{LMLP}
\newcommand{\clutrr}[0]{CLUTRR-LP}
\newcommand{\countries}[0]{Countries-LP}

\newcommand{\R}{\mathcal{R}}
\newcommand{\F}{\mathcal{F}}

\usepackage{pgffor}
\usepackage{adjustbox}
\usepackage{url}
\usepackage{wrapfig}

\usepackage{amsmath}
\usepackage{amssymb}
\usepackage{mathtools}
\usepackage{amsthm}

\usepackage[capitalize,noabbrev]{cleveref}

\theoremstyle{plain}

\theoremstyle{definition}

\theoremstyle{remark}

\usepackage[textsize=tiny]{todonotes}

\usepackage{booktabs}
\usepackage{microtype} %
\usepackage[bottom]{footmisc}
\usepackage{caption}
\usepackage{helvet}  %
\usepackage{courier}  %
\usepackage{url}  %
\usepackage{graphicx}  %
\usepackage{multirow}
\usepackage{amsthm}
\usepackage{color}
\usepackage{MnSymbol}
\usepackage{arydshln}
\usepackage{amsmath}
\usepackage{xcolor}
\usepackage{caption} 
\usepackage{natbib}
\usepackage{textcomp}
\usepackage{wrapfig}
\usepackage{algorithm}
\usepackage{algpseudocode}
\usepackage{subfigure}
\usepackage{makecell}
\usepackage{csquotes}
\usepackage{bbding}
\usepackage{utfsym}

\newcommand{\Task}{\textcolor{cyan}{Task}}
\newcommand{\Story}{\textcolor{cyan}{Story}}

\newcommand{\wrong}{\textcolor{red}{\usym{2717}}}
\newcommand{\coloright}{\textcolor{green}{\usym{2713}}}

\usepackage[most]{tcolorbox}

\definecolor{cyan}{HTML}{5785CD}
\definecolor{orange}{HTML}{FF9908}

\definecolor{scllightgrey}{rgb}{0.94,0.94,0.94}%
\definecolor{sclgreyblue}{rgb}{0.2,0.3,0.5}%
\newcommand{\sclcode}[1]{{\tcbox[
on line, capture=hbox, width=.47\textwidth,
boxsep=0pt,left=1pt,right=1pt,top=1pt,bottom=1pt,colframe=scllightgrey,colback=scllightgrey
]{\color{sclgreyblue} \small \texttt{#1}}}}

\title{Evaluating Step-by-Step Reasoning through Symbolic Verification}

\author{ \ %
  Yi-Fan Zhang$^{1}$ \quad
  Hanlin Zhang$^{2}$ \quad %
  Li Erran Li$^{3}$\thanks{~~Work done outside of Amazon.} \quad %
  Eric Xing$^{1,2,4}$ \\
  $^1$MBZUAI
  $^2$Carnegie Mellon University, 
  $^3$AWS AI, Amazon,
  $^4$Petuum Inc.  
  \\
}

\begin{document}
\maketitle

\begin{abstract}

Pre-trained language models (LMs) have shown remarkable reasoning performance using explanations or chain-of-thoughts (CoT)) for in-context learning. 
On the other hand, these reasoning tasks are usually presumed to be more approachable for symbolic programming. 
To understand the mechanism of reasoning of LMs, we curate synthetic datasets containing equivalent (natural, symbolic) data pairs, where symbolic examples contain first-order logic rules and predicates from non-parametric knowledge bases (KBs), supporting automated verification of intermediate reasoning results.
Then we revisit neuro-symbolic approaches and propose to learn from demonstrations containing logic rules and corresponding examples to iteratively reason over KBs, recovering Prolog’s backward chaining algorithm and supporting automated verification of LMs' outputs. 
Comprehensive experiments are included to systematically compare LMLP with CoT in deductive reasoning settings, showing that LMLP enjoys more than $25\%$ higher accuracy than CoT on length generalization benchmarks even with smaller model sizes.

\end{abstract}

\section{Introduction}

There are emerging interests in leveraging LMs to enable planning \citep{li2022pretrained, huang2022language}, heuristic search \citep{dahlgren2021perception} and symbolic inference \citep{wei2022chain, zelikman2022star, zhang2022improved}. 
Among them, \textit{chain of thought} prompting or scratchpads \citep{wei2022chain, nye2021show} shows that taking \sclcode{(input, explanation, output)} as in-context examples for LMs can lead to significant performance gain in reasoning tasks. 
However, like many fine-tuning approaches, it can be difficult for these models to generalize compositionally \citep{zhou2022least}, meaning they may struggle to apply their knowledge to solve new problems that involve novel combinations of information \citep{lake2018generalization, bahdanau2018systematic, keysers2019measuring}.
\begin{figure}[t]
    \centering
    \includegraphics[width=0.5\textwidth]{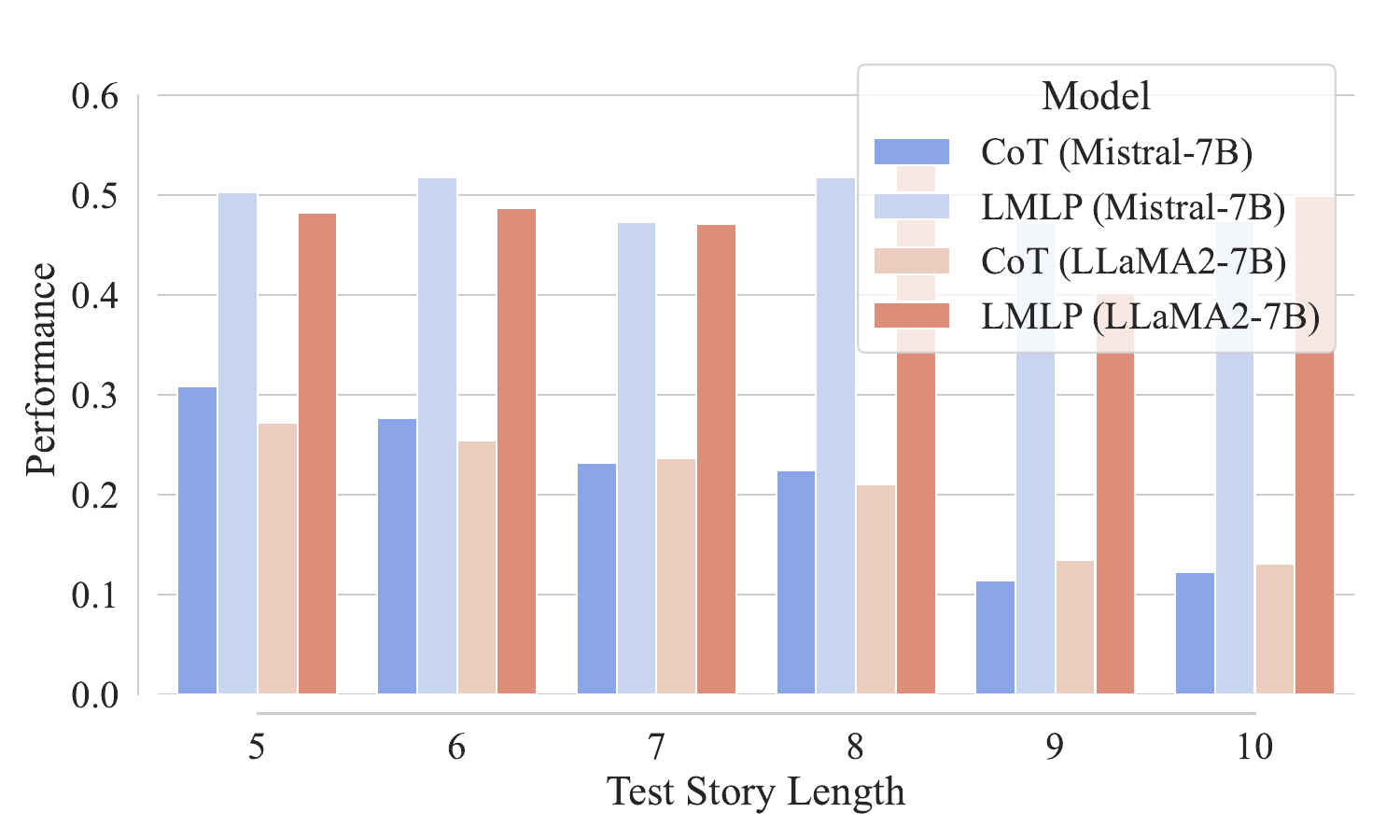}
    \caption{Deductive reasoning performance (human evaluation accuracy) comparisons on the \clutrr~ given training data with story length 2, 3, 4. }
    \label{fig:lm_cot}
    \vspace{-0.1cm}
\end{figure}
One notable case is that LMs would suffer from catastrophic performance degradation when tested on sequences longer than training ones (Figure~\ref{fig:lm_cot}). As a solution, \textit{least-to-most prompting} \citep{zhou2022least} takes inspiration from symbolic programs and proposes to tackle the challenge by modularizing the prompt on the reduced problem. 
The divide-and-conquer strategy is useful to improve the reasoning ability of language models, but it also presents additional challenges: 
what are the appropriate representations for factual knowledge and in-context samples that can ensure the correctness of each individual reasoning step? 
How do natural language explanations compare to symbolic provenance, which is easily verifiable, when used as prompts for reasoning?

Our goal is to evaluate the natural and symbolic paradigms closely in order to answer these questions. To enable fine-grained comparison and gain insight into in-context learning for reasoning tasks, we study relational reasoning over both natural language and knowledge bases (KBs). 
KBs are particularly useful for this purpose because they are constructed using clear pipelines and strong supervision, which makes them reliable and easy to control. This allows us to verify and evaluate reasoning paths and provenances without the need for human-provided rationales or explanations \citep{camburu2018snli, zhou2020towards, wei2022chain, nye2021show, zelikman2022star}.
We study language models as logic programmers (\abbr) to enable few-shot learning from symbolic demonstrations and simultaneous planning in an explainable and scalable way. \abbr~ uses logic rule templates, examples, and pre-trained knowledge to iteratively perform in-context learning and answer relational queries.

\begin{figure}[t]
    \centering
    \includegraphics[width=0.49\textwidth]{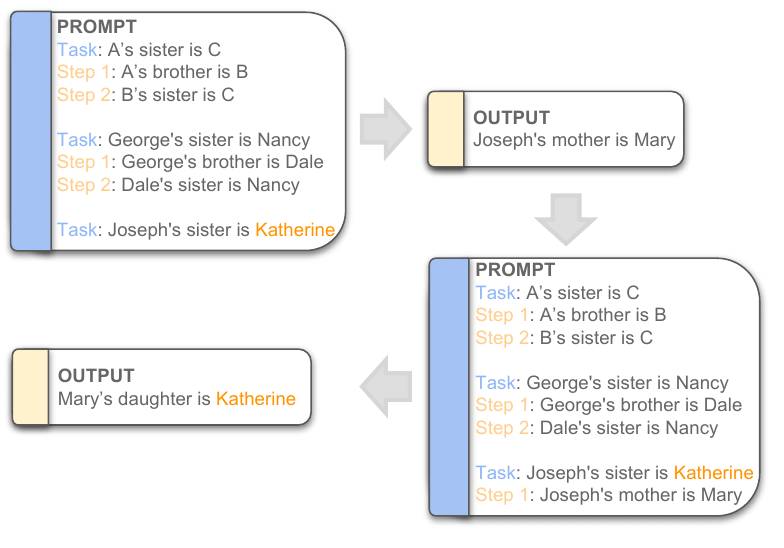}
    \caption{Illustration of a deductive reasoning example and iterative prompting of LMLP. 
    LMLP retrieves a first-order logic rule and an associated grounded example to answer the question. 
    It stops when predefined maximum iterations or the \textcolor{orange}{target} entity of interest is reached. The reasoning path explains the \textit{sister} concept.}
    \label{fig:process}
\end{figure}

Specifically, given a goal query as the in-context example that can be interpreted as a question answering (QA) task, \abbr~ searches or retrieves a related task example with a corresponding logic rule (Figure~\ref{fig:process}). Then the context and task description are concatenated as the input prompt for an autoregressive planning LM. 
At each step of generation, we use a masked translation LM to compare the similarity between the generated natural language sentences and encoded it into \sclcode{(subject, relation, object)} predicates in the KB. 
In this way, each generated sentence is transformed into the most similar predicate and the reasoning path is confined within the KB. 
The process is iterated until a predefined maximum iteration or the target of interest is reached (Figure~\ref{fig:process}) and the generated reasoning path is evaluated manually.

To evaluate the reasoning capability of CoT and LMLP, we curate two datasets and design a series of experiments, aiming to compare two recent in-context learning paradigms and explore both symbolic and naturalistic scenarios. 
Specifically, we adopt synthetic datasets containing \sclcode{(natural, symbolic)} data pairs. 
The symbolic part contains predicates and first-order logic (FOL) rules, which are well-suited for investigating the role of symbolic representations for few-shot reasoning. 
The natural part of our study includes a story written in natural language that describes a set of entities and relations, as well as the reasoning paths that connect them. These reasoning paths can be seen as explanations for the relationships and events described in the story.
Moreover, we create experimental settings that are unfavorable for LMLP since 
(i) we use GPT-2 and SentenceBERT as its backbones, which is known to be of much smaller scale compared to CoT which is usually based on GPT-3 \citep{brown2020language} or PaLM \citep{schowdhery2022palm}; 
(ii) LMs are pre-trained over natural language sentences as opposed to KBs, which creates substantial gaps in semantics and representations, thus posing a grounding challenge where LMs are known to be ineffective \citep{bisk2020experience}.

Controlled experiments on relational reasoning have shown that 
(i) CoT prompting struggles to solve the compositionality challenge \citep{sinha2019clutrr}, while with explicit verification, LMLP can work more reliably as reasoning length increases by taking symbolic inputs that explicitly separate logic and control \citep{kowalski1979algorithm}. 
(ii) While it is commonly believed that large pre-trained language models (LMs) are not grounded in contexts that require rich experiences, experimental results suggest that in-context learning, which maps the conceptual structure of a space learned from text onto a new structured space, is sufficient to solve some challenging reasoning tasks over knowledge bases (KBs). 
(iii) LMs struggle to effectively solve relational reasoning tasks without proper demonstrations containing the target relation and correct input-label mappings. This is supported by evidence in in-context examples, which are poorly understood and have many intricate design choices \citep{zhao2021calibrate, liu2021makes, min2022rethinking}.

\vspace{-0.1cm}
\section{Related Works}
\vspace{-0.1cm}
\textbf{In-context learning} concerns feeding input texts describing a task with some examples to the black-box model for learning the task \citep{brown2020language}. Many works show that there are intricate design choices like prompt formats \citep{jiang2020can, liu2021makes, zhao2021calibrate, min2022rethinking}, example choices and their ordering \citep{zhao2021calibrate, lu2021fantastically}, pretraining data distribution \citep{xie2021explanation, shin2022effect, chan2022data} and model architectures \citep{chan2022data} to improve the LMs' powerful and versatile in-context learning ability. 
Recent work focuses on bootstrapping LM with natural language explanations, intermediate steps, or rationales for reasoning \citep{camburu2018snli, zhou2020towards, nye2021show, wei2022chain, nye2021show, zelikman2022star}. 
Recent works showcase both some positive \citep{clark2021transformers} and negative results \citep{kassner2020pretrained, helwe2021reasoning, talmor2020olmpics} in adapting LMs for symbolic or logical reasoning. 
The length generalization challenge is echoed in a few recent works  
\citep{zhang2022improved, anil2022exploring, liu2022transformers, zhou2022teaching, press2022measuring}.
Though there are some encouraging progress \citep{clark2021transformers, wei2022chain, schowdhery2022palm, zelikman2022star}, they require a significant amount of computation for re-training and human annotations about reasoning paths or explanations \citep{wei2022chain, nye2021show}. Moreover, their entangled nature with natural language makes them hard to make robust inferences over symbolic factual knowledge. 
However, our goal is fundamentally different from theirs in investigating the role of symbolic representations on few-shot reasoning using in-context learning. 
\abbr~ that bootstraps the reasoning process from the LMs in a few-shot manner (Figure \ref{fig:process}) is in contrast to popular methods that need expensive human annotations and retraining \citep{camburu2018snli, zhou2020towards, wei2022chain, zelikman2022star} or uncontrollable using only pre-trained knowledge~\citep{zero2022kojima}. 
Moreover, related works typically \textit{finetune} the model using rationales or explanations \citep{camburu2018snli, zhou2020towards} or focus on natural language based reasoning such as commonsense reasoning, arithmetic reasoning, open domain question answering \citep{wei2022chain}, concept grounding \citep{patel2021mapping} etc. 
Synthetic ontology datasets are constructed in \citep{saparov2022language} to understand the failure modes of CoT reasoning, but they are in natural language forms instead of investigating the reasoning done over interpretable symbolic structures as we do. \citet{huang2022language} uses a mechanism for constraining the LLM output to feasible action sequences, which we adopt in this work. \abbr~ can be conceptually understood as a realization of recency biases \citep{press2021train}, which has been shown effective in scratchpad-based reasoning \citep{liu2022transformers}.
Therefore, all the above works are different from our goal of exploring the representations of prompts in-context learning. 

\textbf{Retrieval-augmented Generation.} Our study is also related to retrieval-augmented generation \citep{lewis2020retrieval} like kNN-LM \citep{khandelwal2019generalization}, DPR \citep{karpukhin2020dense}, RALM \citep{guu2020retrieval}, and RETRO \citep{borgeaud2022improving}, which integrates parametric models with non-parametric KBs to address key LM challenges like knowledge staleness \citep{roberts2020much} and hallucination \citep{shuster2021retrieval}, reasoning \citep{shao2023enhancing}. 
We explore more controllable environments where the evaluation of intermediate reasoning can be automated, demonstrating that this verification process helps filter out incorrect reasoning paths. This, in turn, enhances reasoning performance by assessing how effectively language models can reason when instances of hallucination are minimized.
\vspace{-0.08cm}

 \vspace{-3mm}

\section{Methodology Overview}
\vspace{-2mm}
We consider the reasoning task with an SRL query as the question and some background knowledge as the context.
The relational information in the query and context can be expressed either using natural language or a \sclcode{(subject, relation, object)} predicate/triplet.
There is a KB with facts $\F$ and (FOL) rules $\R$ to support the above QA. 
There are two equivalent ways for representing the problem, symbolic or natural language, which leads to the designs below.

\textbf{Datasets construction.} \label{sec:data_process}
To ensure that the natural and symbolic data are equivalent, we keep the ground truth facts the same in natural language stories and knowledge bases. We construct natural language story datasets following the method described in \citep{sinha2019clutrr}.
As shown in Table~\ref{tab:data_process}, we seek to curate new symbolic datasets from the original ones into 
(i) \textit{A query subset} containing predicates needed for proving. 
(ii) \textit{A set of facts $\F$} containing all the available facts/predicates, which composes a KB, and 
(iii) \textit{A set of rules $\R$} containing examples (A task and its proofs) extracted from the training subset using backward chaining based neuro-symbolic reasoners~\citep{rocktaschel2017end}. 
See appendix~\ref{app:data} for more details.

\begin{figure*}[htbp]
\centering
\subfigure[LMLP]{
\begin{minipage}[t]{0.45\linewidth}
\centering
\includegraphics[width=\textwidth]{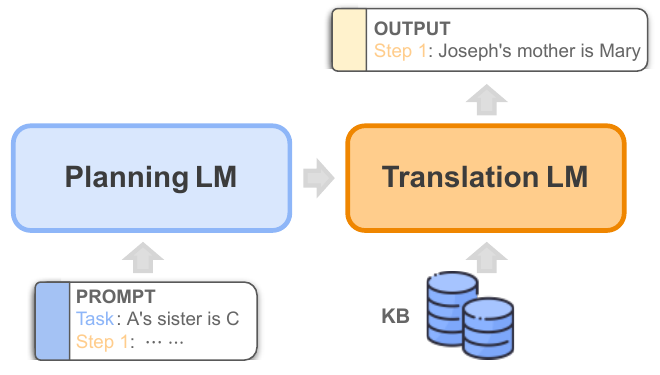}
\label{fig:model}
\end{minipage}%
}%
\subfigure[CoT]{
\begin{minipage}[t]{0.49\linewidth}
\centering
\includegraphics[width=\textwidth]{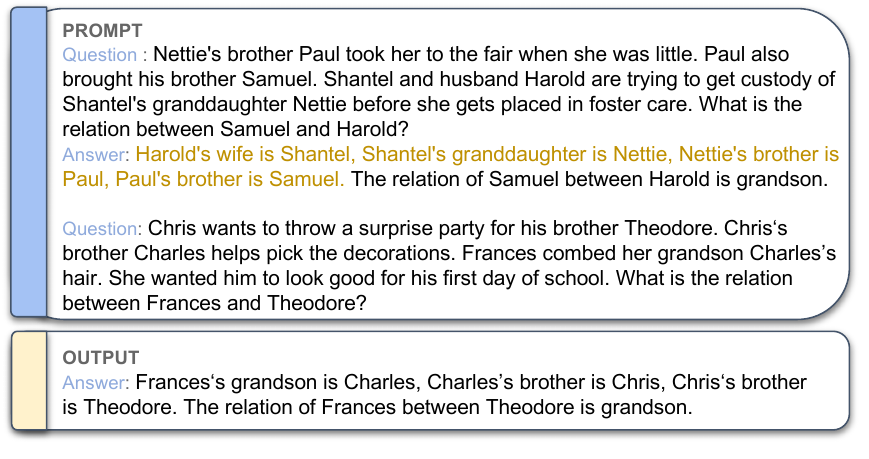}
\label{fig:model_cot}
\end{minipage}%
}%
\centering
\vspace{-0.1cm}
\caption{Schematic overview of (a) \abbr~and (b) CoT.} 
\vspace{-0.1cm}
\end{figure*}

\begin{table*}[h]
\adjustbox{max width=\textwidth}{%
\begin{tabular}{cclcl}
\hline
\textbf{Dataset} & \textbf{Natural Language Samples} & \multicolumn{1}{c}{\textbf{Query}} & \textbf{Facts $\mathcal{F}$} %
& \multicolumn{1}{c}{\textbf{Logic rules $\mathcal{R}$}} %
\\ \hline
\textbf{CLUTRR} &  \multicolumn{1}{l}{
\begin{tabular}[c]{@{}l@{}}
\Task: What's the relationship between Ashley and Nicholas? \\ 
\Story: Ashley told her daughter Lillian to wash up.\\  
 Dinner was ready. Lillian called her brother, Nicholas \\  up to see how he was doing after surgery.
\end{tabular}} 
& (Ashley, son, Nicholas)             
& \multicolumn{1}{l}{
\begin{tabular}[c]{@{}c@{}}
(Ashley, daughter, Lillian) \\ 
(Lillian, brother, Nicholas) \\ \centering{$\cdots$}
\end{tabular}} & 
\begin{tabular}[c]{@{}l@{}}
\Task: Ashley's son is Nicholas \\ 
\textcolor{orange}{Step 1}: Ashley's daughter is Lillian \\ \textcolor{orange}{Step 2}: Lillian's brother is Nicholas\end{tabular} \\ \hline
\textbf{Countries} & 
\begin{tabular}[c]{@{}l@{}} 
\Task: Is palau located in oceania? %
\end{tabular}
& (palau, locatedIn, oceania)             & \multicolumn{1}{l}{
\begin{tabular}[c]{@{}c@{}}(palau, locatedIn, micronesia)\\ \centering{$\cdots$}
\end{tabular}}   & \begin{tabular}[c]{@{}l@{}}\Task: palau locatedIn oceania \\
\textcolor{orange}{Step 1}: palau locatedIn micronesia \\ \textcolor{orange}{Step 2}: micronesia locatedIn oceania\end{tabular}   \\ \hline
\end{tabular}}
\caption{Examples of data processing and curation.}
\vspace{-0.1cm}
\label{tab:data_process}
\end{table*}
\textbf{Task.} Given a query \textit{Task: Joseph's sister is Katherine}, which consists of two entities \sclcode{Joseph, Katherine} and a target {relation} \sclcode{sisiter}. 
Our task is to find a proof path from \sclcode{Joseph} to \sclcode{Katherine} where the relationship \sclcode{sisiter} can be correctly inferred. 
On a high level, we need to leverage an abstract logic rule and its grounded example:
\sclcode{Sister(A,C) $\leftarrow$ Brother(A,B) $\land$ Sister(B,C)}
\sclcode{Sister(George, Nancy) $\leftarrow$ Brother(George,} \sclcode{Dale)
$\land$ Sister(Dale, Nancy)}
to derive the answer for the query \sclcode{Sister(Joseph, Katherine)} (Figure~\ref{fig:model}).

\textbf{Language Models as Logic Programmers} achieves this goal using in-context learning. At first, examples and logic rules $r$ in $\mathcal{R}$ are selected.
For example, in Figure~\ref{fig:process}, LMLP samples one logic rule and its grounded example, which is concatenated with the query $q$ \sclcode{Task: Joseph's sister is Katherine} as a prompt $r'=[r, q]$. The prompt is fed into a \textbf{Planning LM} $\mathcal{P}_\theta$, which is an autoregressive LM such as GPT-3 for proof generation. Multiple sentences $x$ are generated using temperature sampling from $\mathcal{P}_\theta(r')$. However, these sentences are in free-form language and often not in the \sclcode{(subject, relation, object)} predicate format. In LMLP, the generated output is converted to the most similar fact in KB $\F$ using the cosine similarity of the embedding from a \textbf{Translation LM} $\mathcal{T}_\phi$, implemented as a sentence-specific Masked LM.  
Specifically, $\mathcal{T}_\phi$ embed the output sentence from $\mathcal{P}_\theta$: $\mathcal{T}_\phi(x)$ and all predicates $f$ from $\mathcal{F}$: $\mathcal{T}_\phi(f)$, calculating their cosine similarity. The most similar $f$ to $x$ is chosen as the conversion results $f'$. 
By translating the output space of $\mathcal{P}_\theta$ into an external KB this way, \abbr~is expected to produce a more plausible provenance to explain the reasoning process of a final prediction.
Given frozen $\mathcal{P}_\theta$ and $\mathcal{T}_\phi$, we then repeatedly generate proofs by prompting $\mathcal{P}_\theta$ using $r'=[r',f']$, projecting the generated sentences to the KB by the $\mathcal{T}_\phi$, attaching the output to the prompt (Figure~\ref{fig:process}). 
The model terminates when the predefined maximum number of iterations or the target entity of interest is reached.
To improve coherency, we enforce the chain rule transition constraints: the tail entity of the previous predicate should be the same as the head entity of the next predicate for each output step. Specifically, during the translation phase, we only select the predicates satisfying the requirement to compare similarity with $\mathcal{T}_\phi(x)$.
The faithfulness of the reasoning path is governed by post-hoc human evaluations.
The overall algorithm is described in Algorithm \ref{algo:main} in Appendix \ref{app:algo}.
Using the prompt supported by the KBs, we bootstrap the reasoning process from the LMs in a few-shot manner (Figure \ref{fig:process}). 

\textbf{Chain-of-Thought prompting.} CoT~\citep{wei2022chain} solves complicated multi-step reasoning tasks by providing explanations, which is also intuitive for our multi-hop SRL tasks since we can take intermediate reasoning paths as explanations. 

\figurename~\ref{fig:model_cot} shows an example of applying CoT to solve an SRL task from the CLUTRR dataset \citep{sinha2019clutrr}: 
given an in-context sample in the form of 
\sclcode{(input, \textcolor{brown}{explanation}, output)}.
LMs are expected to imitate the reasoning process of the given explanation to generalize to a new query. 
The explanation of each question is generated just the same as the rule set $\mathcal{R}$, which is extracted from the training set using a neuro-symbolic reasoners and converted to natural language forms.
Specifically, the in-context exemplar adapts LMs to another sample containing multiple relations and a query for the relation between two entities \textit{``What is the relation between Theodore and Frances?''}, 
CoT first generates a reasoning path from \textit{Frances} to \textit{Theodore}, namely \textit{``France's grandson is Charles, \dots, Chris's brother is Theodore.''}, and finally answers the query: \textit{``The relation of Frances between Theodore is grandson''}. 
With such a prompt, LMs are expected to generate both the reasoning paths and the resulting queried relation. 
For a fair comparison with \abbr, human judgments on the reasoning path are included to calculate the accuracy.
Note that the explanation in CoT is extracted from the story in the question, which contains much clearer information than the logic rules for \abbr. 

  \vspace{-3mm}
\section{Experiments}

We now describe the experimental setups, empirically evaluate \abbr~and compare it with existing methods. See Appendix \ref{app:exp} for full details of data preprocessing and performance evaluation.

\textbf{Settings.} We curate two datasets for evaluating the in-context learning capability of LMs for reasoning: \clutrr~and \countries, which are based on CLUTRR \citep{sinha2019clutrr} and Countries~\citep{bouchard2015approximate} datasets respectively. CLUTRR \citep{sinha2019clutrr} contains a group of KBs, where each node denotes a family member and edges are family relations. The target of CLUTRR dataset is to infer a two-family members' relationship that is not explicitly mentioned. 
The training set of CLUTRR consists of graphs that the target relation can be inferred by traversing a limited number of edges while the relation in the test set needs more traversing steps for inference, which allows controlled studies on compositionality. 
Another intriguing property of CLUTRR is that there are ground truth \textbf{one-to-one correspondances} between KBs and natural language stories, which exactly suits our needs.
Countries~\citep{bouchard2015approximate} concerns link prediction, where countries, regions, and sub-regions are entities and relations containing \textit{LocatedIn} and \textit{NeighborOf}. 
Countries has three tasks, \textit{R1,R2}, and \textit{R3}, each requiring reasoning skills of increasing complexity~\citep{rocktaschel2017end}.

\textbf{Implementation details.} 
For \abbr, we implement the planning LM $\mathcal{P}_\theta$ as GPT-2 \citep{radford2019language}, the translation LM $\mathcal{T}_\phi$ as Sentence BERT (Sent-BERT) \citep{reimers2019sentence} based on Hugging Face Transformers~\citep{wolf2019huggingface}. 
The default model for Translation LM is Sentence-RoBERTa-Large and for Planning LM is GPT2-Large \citep{radford2019language} pretrained on large corpora by default. 
For CoT, we follow the original paper \citep{wei2022chain} to sample in-context samples. 

Since prompt formats lead to significant performance variations \citep{liu2021makes}, we propose to explore two simple design choices for \abbr~and find that they can further boost the reasoning capacity.
(i) Multiple examples for prompting. Denote $N$ the number of examples we used in one proof task. Table~\ref{tab:comp_prompt} shows two examples with $N=1$ and $N=2$ are supplied respectively. 
The intuition is that, getting more examples in the prompt can make LMs better recognize the proof task and thus produce more reliable reasoning paths. 
See the experimental section for empirical verification.
(ii) Prompts Ensembling. Table~\ref{tab:example_multi_prompt} shows the results of {different prompts for the same task}. We can see the influence of prompts on the generated proof path. 
The first few proof steps are largely similar to the provided example. If the provided example supplies a wrong direction, the proof is likely to be wrong. 
To study and exploit the benefit brought by different prompts, 
during experiments, we propose to use $K$ prompts alternatively for one task, where one task is marked to be successfully proved if any of these $K$ prompts gets the right result. Namely, a larger $K$ means that we have a higher probability of picking a good prompt. 
The default hyper-parameters $N,K$ are set to one.

\textbf{Evaluation metrics} In Table~\ref{tab:human_evaluation} and Table~\ref{tab:human_countries}, where \abbr~is compared to various baselines, the correctness of the proven reasoning path is evaluated manually. For each reasoning path, we ask annotators to answer ``Yes'' or ``No'' to whether the generated proof path is plausible to human commonsense and the target relation can be induced from it. 
We include 5 participants to reduce randomness and observe that their answers are almost the same. Because of resource Limitations, for other simple ablation studies of \abbr, the metric is \textbf{proven accuracy or success rate}. For example, for query ``Task: palau locatedIn oceania'', we begin with entity ``palau'' and select facts from the $\F$. If the chosen triplet ends with entity ``oceania'', the proven path is correct, e.g., ``micronesia locatedIn oceania'' in Table~\ref{tab:data_process}. For \abbr, if there is no chosen triplet ends with entity ``oceania'', the prediction is incorrect.

\begin{table*}[h]
\centering
\adjustbox{max width=\textwidth}{%
\begin{tabular}{ccc|cccc|cc}
\toprule 
\multirow{2}{*}{Test Story Length} & \multicolumn{2}{c}{Baseline} & \multicolumn{4}{c}{Ablation} & \multicolumn{2}{c}{Ours} \\
 & \textbf{Planner} & \textbf{CoT} & \textbf{No Prompt} & \textbf{Only Rule} & \textbf{Random} & \textbf{Entity-based} & \textbf{\abbr-reverse} & \textbf{\abbr} \\ \hline
5 & 0.0973 & 0.173 & 0.1514 & 0.1622 & 0.2919 & 0.2000 & \textbf{0.3730} & 0.3297 \\
6 & 0.1810 & 0.1365 & 0.1238 & 0.1524 & 0.2095 & 0.1429 & \textbf{0.3048} & 0.2476 \\
7 & 0.2258 & 0.1032 & 0.2000 & 0.2129 & 0.2323 & 0.1742 &\textbf{ 0.3742} & 0.2581 \\
8 & 0.1037 & 0.1506 & 0.2222 & 0.2000 & 0.3111 & 0.2370 & \textbf{0.3556} & \textbf{0.3556} \\
9 & 0.1048 & 0.0914 & 0.1935 & 0.2177 & 0.1613 & 0.1855 & \textbf{0.3548} & 0.2984 \\
10 & 0.1230 & 0.123 & 0.2869 & 0.2131 & 0.3934 & 0.2705 & 0\textbf{.5246} & 0.4754 \\
Average & 0.1393 & 0.1296 & 0.1963 & 0.1931 & 0.2666 & 0.2017 & \textbf{0.3812} & 0.3275\\ \bottomrule
\end{tabular}%
}
\caption{Numerical results and ablation on the length of test samples on \clutrr.}\label{tab:ablation}\label{tab:human_evaluation}
\end{table*}

\begin{table*}[]
\centering
\resizebox{0.9\textwidth}{!}{%
\begin{tabular}{@{}ccccccccc@{}}
\toprule
\multirow{2}{*}{Tasks} & \multicolumn{2}{c}{Baseline} & \multicolumn{4}{c}{Ablation} & \multicolumn{2}{c}{Ours} \\

 & \textbf{Planner} & \textbf{CoT} & \textbf{No Prompt} & \textbf{Only Rule} & \textbf{Random} & \textbf{Entity-based} & \textbf{\abbr-reverse} & \textbf{\abbr} \\ \midrule
S1 & 0.7500 & 0.3333 & 0.8542 & 0.7708 & 0.6042 & \textbf{0.8958} & 0.8333 & 0.7917 \\
S2 & 0.7917 & 0.3750 & 0.6667 & 0.4583 & 0.6750 & 0.7500 & \textbf{0.8333} & 0.6250 \\
S3 & 0.7500 & 0.2500 & 0.7292 & 0.7083 & 0.6458 & 0.6667 & 0.7500 & \textbf{0.8333} \\
Average & 0.7639 & 0.3194 & 0.7500 & 0.6458 & 0.6417 & 0.7708 & \textbf{0.8055} & 0.7500 \\
\bottomrule
\end{tabular}%
}
\caption{Human evaluation results in various settings of \countries. \textit{S1, S2, S3} ~\citep{minervini2020learning} are three different tasks with different $\F$ (see the experimental setting for details).}\label{tab:human_countries}
\end{table*}
\subsection{Comparisons of \abbr~and CoT}%

The goal of this part is to systematically compare \abbr~with CoT both quantitatively and qualitatively on SRL tasks to better understand the reasoning of LMs using in-context learning.

In~\figurename~\ref{fig:lm_cot} and Table.~\ref{tab:result_models}, we compare \abbr~to CoT and the reported performances are all human evaluation results. 
Qualitatively, CoT can get positive results on some query examples, for example, in Table~\ref{tab:example_cot}, we showcase two examples where CoT can generate a correct proof path and predict the target relation at the same time. 
However, compared to \abbr, CoT achieves inferior results in all query sets with test reasoning length $5,6,7,8,9,10$ with different LLMs for text generation. In addition, as the reasoning length increases, the performance of CoT shows a clear downward trend. 
Table~\ref{tab:example_cot} shows two negative examples where the story contains sophisticated relations and the model cannot get the right reasoning path or just generate a wrong relation. 
In contrast, \abbr~can consistently achieves a high human evaluation score (Table~\ref{tab:human_evaluation}), which again verifies the systematic generalization capability of \abbr. Table~\ref{tab:lm_cot} in the appendix shows examples with the same task but processed by the two methods respectively, where CoT cannot get deduce a right relation path from \textit{Margaret} to \textit{Charles} but \abbr~can extract a simple yet right relation path.
The reason why \abbr~is better than CoT can be that, although CoT decomposes complex multi-hop relation reasoning tasks into a multi-step reasoning process and then predict the final results, the proof path is all generated by LMs at once. 
The decomposition of \abbr~to multi-hop reasoning tasks is more thorough, where the generation of a proof path is divided into multi-steps and each step will be projected into the KB, which is a much stronger inductive bias.
Therefore, the decomposed tasks in each step are easier to solve and the knowledge in the KB can be well exploited. See appendix for results on \countries.

\begin{table}[htbp]
  \centering
  \resizebox{\linewidth}{!}{%
  \begin{tabular}{ccccccccc}
    \hline
     \multirow{2}{*}{Test Story Length} & \multicolumn{2}{c}{GPT-2} &   \multicolumn{2}{c}{Mistral-7B-v0.1} &  \multicolumn{2}{c}{ LLaMA2-7B } \\
     & \textbf{CoT} &\textbf{LMLP} & \textbf{CoT} & \textbf{LMLP} & \textbf{CoT} & \textbf{LMLP} \\
    \hline
    5 & 0.1730 & 0.3297 & 0.3083 & 0.5032 & 0.2721 & 0.4823 \\
    6 & 0.1365 & 0.2476 & 0.2762 & 0.5182 & 0.2543 & 0.4872 \\
    7 & 0.1032 & 0.2581 & 0.2314 & 0.4732 & 0.2364 & 0.4715 \\
    8 & 0.1506 & 0.3556 & 0.2247 & 0.5181 & 0.2102 & 0.5323 \\
    9 & 0.0914 & 0.2984 & 0.1143 & 0.4723 & 0.1345 & 0.4021 \\
    10 & 0.1230 & 0.4754 & 0.1220 & 0.4741 & 0.1305 & 0.4992 \\
    \hline
    Average & 0.1296 & \textbf{0.3275} & 0.2128 & \textbf{0.4932} & 0.2063 & \textbf{0.4791} \\
    \hline
  \end{tabular}}
  \caption{Numerical results considering different backbone models.}
  \label{tab:result_models}
\end{table}

\subsection{Analysis of \abbr}

Given the above observations that \abbr~outperforms CoT by a large margin, we systematically analyze \abbr~with extensive experiments below.

\textbf{Ablation Studies on prompting strategies.} %
As illustrated in Table~\ref{tab:ablation}, 
\textbf{No Prompt} means that we only feed the target directly and generate each step, 
prompts in the \textbf{Only Rule} baseline is one proof example with entities replaced by some symbols. We also compare \abbr~to Language 
Planner~\citep{huang2022language}, which first finds the most similar target in the $\R$ and uses such an example as the prompt. \textbf{\abbr-reverse} swaps the position of the abstract logic rule and its grounded example in the prompt of \abbr. For example, in~\figurename~\ref{fig:process}, the in-context prompt of \abbr-reverse will place \textit{Sister(George, Nancy) $\leftarrow$ Brother(George, Dale) $\land$ Sister(Dale, Nancy)} before its abstract logic rule \sclcode{Sister(A,C) $\leftarrow$ Brother(A,B) $\land$ Sister(B,C)}
. Examples for all baselines are shown in Appendix Table~\ref{tab:comp_prompt}.

Table~\ref{tab:ablation} shows that directly applying Language Planner for relational reasoning does not work and using only facts or no prompt attain inferior performance. 
The possible reason for the inferior performance of Planner can be that it finds the example from $\R$ with the most similar task as the prompt, which usually retrieves rules with the same entities of the goal task. 
However, for reasoning tasks over KBs, relation contains much more information of the task than the entity. As shown in Table~\ref{tab:comp_prompt}, for the task ``Patricia's uncle is Donald'', Planner finds the example with task ``David's nephew is Don'', whose following proofs do not make sense for the relation ``uncle''. \abbr~in contrast finds an example whose task has the same relation as the goal predicate, which is more informative. 
\begin{figure*}[htbp]
\centering
\subfigure[]{
\begin{minipage}[t]{0.33\linewidth}
\centering
\includegraphics[width=\linewidth]{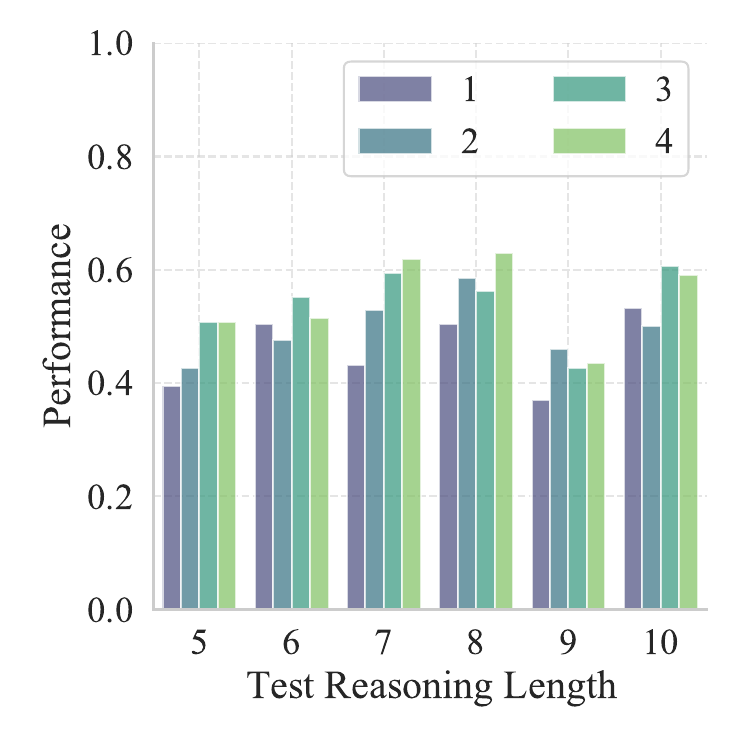}
\label{fig:num_templates_all}
\end{minipage}%
}%
\subfigure[]{
\begin{minipage}[t]{0.33\linewidth}
\centering
\includegraphics[width=\linewidth]{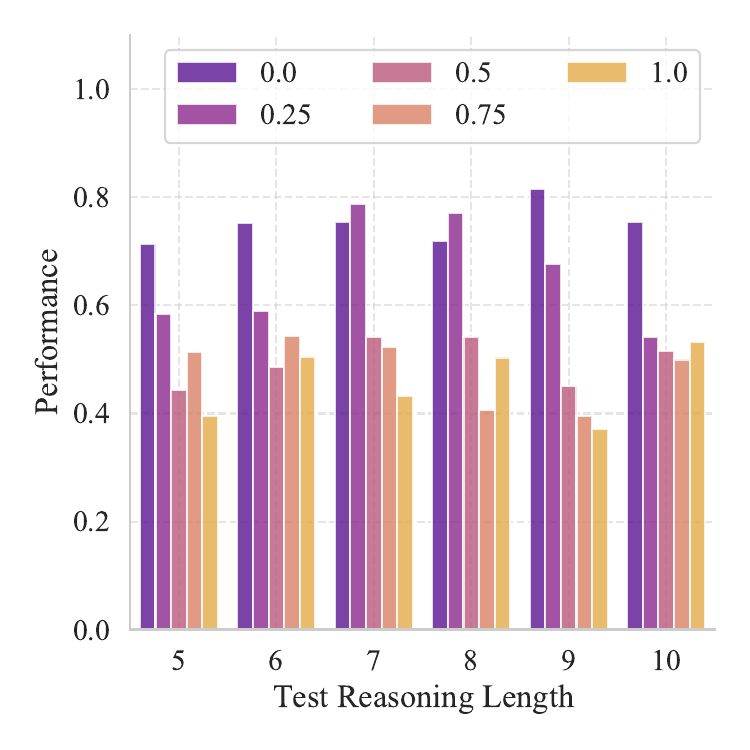}
\label{fig:noisyrate}
\end{minipage}%
}%
\subfigure[]{
\begin{minipage}[t]{0.33\linewidth}
\centering
\includegraphics[width=\linewidth]{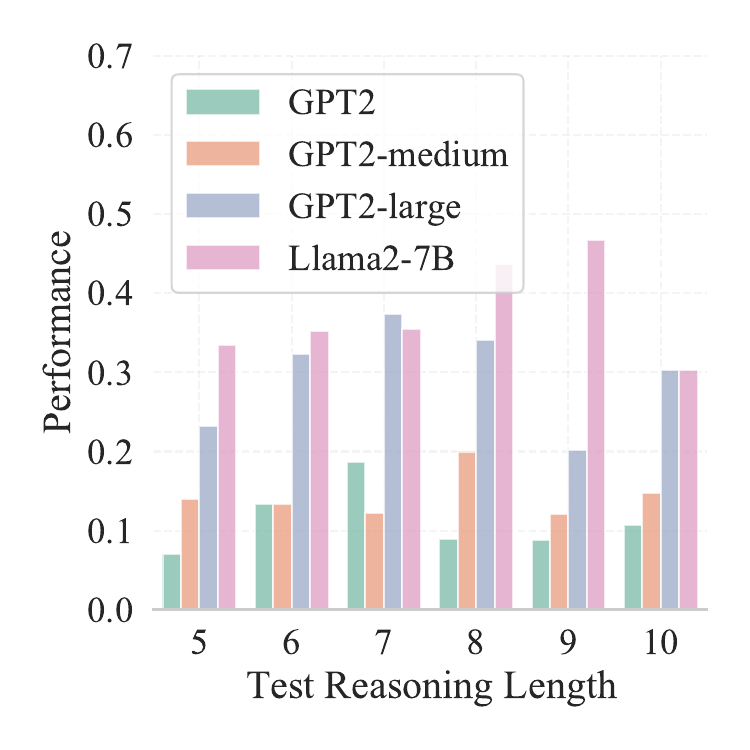}
\label{fig:modelsize_} 
\end{minipage}%
}%
\centering
\vspace{-0.1cm}
\caption{(a) Effect of the number of templates for \abbr~ on \clutrr. (b) The effects of noisy facts for \abbr~ on \clutrr. Ablation on the scaling of (c) Planning LMs.} 
\end{figure*}

\textbf{\abbr~can be robust to large search space.} 
We may wonder if the superior results of \abbr~are an artifact for datasets with a small search space. 
To control the confounding, we progressively inject $5,000$ random noisy facts/predicates into the facts set $\mathcal{F}$. 
With more noisy facts, at each decoding step, it will be more difficult for \abbr~to choose the correct proof path as the search space is enlarged.
\figurename~\ref{fig:noisyrate} shows the results when we vary the number of noisy facts, 
where the noisy rate is $0.5$ means that we add $5000*0.5$ random facts to the $\F$ during evaluation and noisy rate $0$ means $\mathcal{F}$ only contains query-relevant facts. We see that enlarging the search space generally decreases the performance. However, even though when all the noisy facts are injected into $\mathcal{F}$, i.e. more than $95\%$ facts are noisy, 
the performance is still favorable (more than 38\% success rate), showing that \abbr~can produce robust reasoning performance. %

\textbf{Effects of model size.} \figurename~\ref{fig:modelsize_} shows the impact of the size of the planning LM model: larger GPT models generally attain better performance; using GPT2-large and LlaMA2-7B~\citep{touvron2023llama} can dramatically improve model performance, which aligns with the findings that reasoning performance can emerge in larger models \citep{wei2022emergent, saparov2022language}. %

\begin{table}[h]
\centering
\resizebox{\columnwidth}{!}{
\begin{tabular}{@{}cccccl@{}}
\toprule
   & K=1    & K=3    & K=5    & K=10   & \multicolumn{1}{c}{A Long Example}                                                                                                                    \\ \midrule
S1 & 0.7083 & 0.9583 & 1.0000 & 1.0000 & {\begin{tabular}[c]{@{}l@{}}\Task: A locatedIn C\\  \textcolor{orange}{Step 1}: A neighborOf B\\  \textcolor{orange}{Step 2}: B locatedIn C\end{tabular}} \\ \hline
S2 & 0.5000 & 0.8750 & 0.9583 & 1.0000 &       {\begin{tabular}[c]{@{}l@{}}\Task: uruguay locatedIn south\_america\\  \textcolor{orange}{Step 1}: uruguay neighborOf argentina\\  \textcolor{orange}{Step 2}: argentina locatedIn south\_america\end{tabular}}                                                                                           \\ \hline
S3 & 0.7500 & 0.9167 & 0.9167 & 1.0000 &      {\begin{tabular}[c]{@{}l@{}}\Task: sudan locatedIn africa\\ \textcolor{orange}{Step 1}: sudan neighborOf central african republic\\ \textcolor{orange}{Step 2}: central african republic neighborOf chad\\ \textcolor{orange}{Step 3}: chad neighborOf south sudan\\ \textcolor{orange}{Step 4}: south sudan neighborOf dr congo \\
\textcolor{orange}{Step 5}: dr congo neighborOf republic of the congo\\
\textcolor{orange}{Step 6}: republic of the congo locatedIn middle africa\\
\textcolor{orange}{Step 7}: middle africa locatedIn africa\end{tabular}}                                                                                             \\ \bottomrule
\end{tabular}}
\caption{Results of \abbr~on \countries. \textit{S1, S2, S3} ~\citep{minervini2020learning} are three different tasks with different $\F$ (see the experimental setting for details).}
\label{tab:result_countries}
\end{table}

\textbf{Prompts ensembling boosts the reasoning capability.} For each test example, we sample $K$ in-context examples and count as correct if any one of them can solve the task. 
We show the evaluation results on \clutrr~in Table~\ref{tab:clu} and the proposed method can generate realistic and correct proof paths. 
A large $K$ can further boost performance, which also verifies the importance of prompt ensembling: 
Table~\ref{tab:result_countries} shows the performance on \countries~where almost all the query samples can be proved correctly with a large $K$. 
One interesting phenomenon is that \abbr~can generate a much longer proof path even though the proof path length in the rule set $\R$ is less than 3. This manifests a potential improvement with respect to the significant weakness in systematic generalization of fine-tuning or re-training of LMs \citep{sinha2019clutrr}.
The $\R$ of \clutrr~contains only examples whose proof paths are less than five. However, during testing, our model can produce proof paths much longer than five steps and perform well on all query sets.

\textbf{Prompting using multiple examples boosts the reasoning capability.} $N$ denotes the number of in-context examples used in one proof task. Results show that a larger $N$ can generally produce performance gains (\figurename~\ref{fig:num_templates_all}). However, longer prompts require larger GPU memories, so there is a trade-off between memory and performance.
\begin{table}[t]
\centering
\resizebox{.49\textwidth}{!}{
\begin{tabular}{@{}cccccc@{}}
\toprule
Test Reasoning Length &  K=1    & K=3    & K=5    & K=10   & Avg    \\
\midrule
5 Hops            & 0.3946 & 0.6865 & 0.7838 & 1.0000 & 0.7162 \\
6 Hops            & 0.5048 & 0.7143 & 0.7619 & 1.0000 & 0.7452 \\
7 Hops            & 0.4323 & 0.8065 & 0.8774 & 1.0000 & 0.7790 \\
8 Hops            & 0.5037 & 0.8000 & 0.8593 & 1.0000 & 0.7907 \\
9 Hops            & 0.3710 & 0.6452 & 0.7500 & 1.0000 & 0.6915 \\
10 Hops           & 0.5328 & 0.8279 & 0.8525 & 0.9180 & 0.7828 \\ \bottomrule
\end{tabular} }
\caption{Ablation of \abbr~on \clutrr.}
\label{tab:clu}
\end{table}

\subsection{Analysis of Demonstrations of ICL}
Besides results in Appendix Table~\ref{tab:lm_cot}, we conduct qualitative analysis of demonstrations of in-context learning.

\textbf{Failure cases analysis of baselines.} Since the generated sentences are closely related to the prompt, Table~\ref{tab:failure_baselines} in Appendix shows that if we randomly choose prompts, the generated proof path has relations similar to the prompt, but is wrong for the given task. 
For entity-based prompts, since the task has the same start entity as the in-context exemplar, the generated steps 1 in this setting are very similar, leading to many wrong proof paths. Language Planner, without chain rule constraint, the generated triplets are chaos, e.g., in Example 1, the generated proof does even not contain the 
subject ``Jon'' and thus exactly wrong. Although the proposed \abbr~attains a high success rate, there are also some failure cases. As shown in Appendix Table~\ref{tab:example_multi_prompt}, an appropriate prompt needs to be chosen for the right proof paths.

\textbf{Takeaways.} 
Similar to previous work \citep{liu2021makes, min2022rethinking}, we find that in-context learning performance varies greatly with choices of exemplars (Table~\ref{tab:clu}).
One of the key findings in \citep{min2022rethinking} is that even without any labeled data, LMs can achieve k-shot performance by simply prompting with demonstrations containing unlabeled inputs. 
Our findings are generally in-line is in line with the importance of input-label formats highlighted in the work. 
However, we show in Table~\ref{tab:sensitivity} and \ref{tab:comp_prompt} that the correct mapping of rule-example pairs is important since giving only rules with symbols like $X, Y, Z$ rather than concrete entities like $China$ makes \abbr~fail catastrophically. 

\vspace{-2mm}
\section{Concluding Remarks}
\vspace{-2.5mm}

In this study, we systematically examine in-context learning of language models (LMs) from a symbolic reasoning perspective, demonstrating that LMs can be prompted with logical demonstrations to generate plausible explanations for reasoning tasks over knowledge bases (KBs). 
Our evaluation results show that constraining outputs of LMs and ensuring intermediate reasoning correctness are important for reasoning performance, providing new insights into in-context learning and a mechanism to reduce incorrect reasoning through symbolic verification.

\newpage
\section*{Limitations}
Like previous works, we study reasoning empirically without theoretical justifications and focus specifically on synthetic data.
Therefore, our results serve as a proof of concept on investigating how ensuring and reducing hallucination can improve overall reasoning, and might not transfer to more complex reasoning tasks.
Moreover, due to access and computation restrictions, we are not able to conduct experiments with the latest LMs like PaLM \citep{chowdhery2022palm}.

\bibliography{ref.bib}

\begin{thebibliography}{74}
\expandafter\ifx\csname natexlab\endcsname\relax\def\natexlab#1{#1}\fi

\bibitem[{Anil et~al.(2022)Anil, Wu, Andreassen, Lewkowycz, Misra, Ramasesh, Slone, Gur-Ari, Dyer, and Neyshabur}]{anil2022exploring}
Cem Anil, Yuhuai Wu, Anders Andreassen, Aitor Lewkowycz, Vedant Misra, Vinay Ramasesh, Ambrose Slone, Guy Gur-Ari, Ethan Dyer, and Behnam Neyshabur. 2022.
\newblock Exploring length generalization in large language models.
\newblock \emph{NeurIPS}.

\bibitem[{Bahdanau et~al.(2018)Bahdanau, Murty, Noukhovitch, Nguyen, de~Vries, and Courville}]{bahdanau2018systematic}
Dzmitry Bahdanau, Shikhar Murty, Michael Noukhovitch, Thien~Huu Nguyen, Harm de~Vries, and Aaron Courville. 2018.
\newblock Systematic generalization: What is required and can it be learned?
\newblock In \emph{ICLR}.

\bibitem[{Bender and Koller(2020)}]{bender2020climbing}
Emily~M Bender and Alexander Koller. 2020.
\newblock Climbing towards nlu: On meaning, form, and understanding in the age of data.
\newblock In \emph{ACL}.

\bibitem[{Bisk et~al.(2020)Bisk, Holtzman, Thomason, Andreas, Bengio, Chai, Lapata, Lazaridou, May, Nisnevich et~al.}]{bisk2020experience}
Yonatan Bisk, Ari Holtzman, Jesse Thomason, Jacob Andreas, Yoshua Bengio, Joyce Chai, Mirella Lapata, Angeliki Lazaridou, Jonathan May, Aleksandr Nisnevich, et~al. 2020.
\newblock Experience grounds language.
\newblock In \emph{EMNLP}.

\bibitem[{Borgeaud et~al.(2022)Borgeaud, Mensch, Hoffmann, Cai, Rutherford, Millican, Van Den~Driessche, Lespiau, Damoc, Clark et~al.}]{borgeaud2022improving}
Sebastian Borgeaud, Arthur Mensch, Jordan Hoffmann, Trevor Cai, Eliza Rutherford, Katie Millican, George~Bm Van Den~Driessche, Jean-Baptiste Lespiau, Bogdan Damoc, Aidan Clark, et~al. 2022.
\newblock Improving language models by retrieving from trillions of tokens.
\newblock In \emph{International conference on machine learning}, pages 2206--2240. PMLR.

\bibitem[{Bouchard et~al.(2015)Bouchard, Singh, and Trouillon}]{bouchard2015approximate}
Guillaume Bouchard, Sameer Singh, and Theo Trouillon. 2015.
\newblock On approximate reasoning capabilities of low-rank vector spaces.
\newblock In \emph{AAAI}.

\bibitem[{Bouraoui et~al.(2020)Bouraoui, Camacho-Collados, and Schockaert}]{bouraoui2020inducing}
Zied Bouraoui, Jose Camacho-Collados, and Steven Schockaert. 2020.
\newblock Inducing relational knowledge from bert.
\newblock In \emph{AAAI}.

\bibitem[{Brown et~al.(2020)Brown, Mann, Ryder, Subbiah, Kaplan, Dhariwal, Neelakantan, Shyam, Sastry, Askell et~al.}]{brown2020language}
Tom Brown, Benjamin Mann, Nick Ryder, Melanie Subbiah, Jared~D Kaplan, Prafulla Dhariwal, Arvind Neelakantan, Pranav Shyam, Girish Sastry, Amanda Askell, et~al. 2020.
\newblock Language models are few-shot learners.
\newblock \emph{NeurIPS}.

\bibitem[{Camburu et~al.(2018)Camburu, Rockt{\"a}schel, Lukasiewicz, and Blunsom}]{camburu2018snli}
Oana-Maria Camburu, Tim Rockt{\"a}schel, Thomas Lukasiewicz, and Phil Blunsom. 2018.
\newblock e-snli: Natural language inference with natural language explanations.
\newblock \emph{NeurIPS}.

\bibitem[{Chan et~al.(2022)Chan, Santoro, Lampinen, Wang, Singh, Richemond, McClelland, and Hill}]{chan2022data}
Stephanie~CY Chan, Adam Santoro, Andrew~K Lampinen, Jane~X Wang, Aaditya Singh, Pierre~H Richemond, Jay McClelland, and Felix Hill. 2022.
\newblock Data distributional properties drive emergent few-shot learning in transformers.
\newblock \emph{arXiv preprint arXiv:2205.05055}.

\bibitem[{Chaudhuri et~al.(2021)Chaudhuri, Ellis, Polozov, Singh, Solar-Lezama, Yue et~al.}]{chaudhuri2021neurosymbolic}
Swarat Chaudhuri, Kevin Ellis, Oleksandr Polozov, Rishabh Singh, Armando Solar-Lezama, Yisong Yue, et~al. 2021.
\newblock \emph{Neurosymbolic Programming}.
\newblock Now Publishers.

\bibitem[{Chowdhery et~al.(2022{\natexlab{a}})Chowdhery, Narang, Devlin, Bosma, Mishra, Roberts, Barham, Chung, Sutton, Gehrmann et~al.}]{schowdhery2022palm}
Aakanksha Chowdhery, Sharan Narang, Jacob Devlin, Maarten Bosma, Gaurav Mishra, Adam Roberts, Paul Barham, Hyung~Won Chung, Charles Sutton, Sebastian Gehrmann, et~al. 2022{\natexlab{a}}.
\newblock Palm: Scaling language modeling with pathways.
\newblock \emph{arXiv preprint arXiv:2204.02311}.

\bibitem[{Chowdhery et~al.(2022{\natexlab{b}})Chowdhery, Narang, Devlin, Bosma, Mishra, Roberts, Barham, Chung, Sutton, Gehrmann et~al.}]{chowdhery2022palm}
Aakanksha Chowdhery, Sharan Narang, Jacob Devlin, Maarten Bosma, Gaurav Mishra, Adam Roberts, Paul Barham, Hyung~Won Chung, Charles Sutton, Sebastian Gehrmann, et~al. 2022{\natexlab{b}}.
\newblock Palm: Scaling language modeling with pathways.
\newblock \emph{arXiv preprint arXiv:2204.02311}.

\bibitem[{Clark et~al.(2021)Clark, Tafjord, and Richardson}]{clark2021transformers}
Peter Clark, Oyvind Tafjord, and Kyle Richardson. 2021.
\newblock Transformers as soft reasoners over language.
\newblock In \emph{IJCAI}.

\bibitem[{Dahlgren et~al.(2021)Dahlgren, Bj{\"o}rklund, and Drewes}]{dahlgren2021perception}
Adam Dahlgren, Johanna Bj{\"o}rklund, and Frank Drewes. 2021.
\newblock Perception, memory, and inference: The trinity of machine learning.
\newblock In \emph{Is Neuro-Symbolic SOTA still a myth for Natural Language Inference? The first workshop}.

\bibitem[{Glenberg and Kaschak(2002)}]{glenberg2002grounding}
Arthur~M Glenberg and Michael~P Kaschak. 2002.
\newblock Grounding language in action.
\newblock \emph{Psychonomic bulletin \& review}.

\bibitem[{Gontier et~al.(2020)Gontier, Sinha, Reddy, and Pal}]{gontier2020measuring}
Nicolas Gontier, Koustuv Sinha, Siva Reddy, and Chris Pal. 2020.
\newblock Measuring systematic generalization in neural proof generation with transformers.
\newblock \emph{NeurIPS}.

\bibitem[{Guu et~al.(2020)Guu, Lee, Tung, Pasupat, and Chang}]{guu2020retrieval}
Kelvin Guu, Kenton Lee, Zora Tung, Panupong Pasupat, and Mingwei Chang. 2020.
\newblock Retrieval augmented language model pre-training.
\newblock In \emph{International conference on machine learning}, pages 3929--3938. PMLR.

\bibitem[{Helwe et~al.(2021)Helwe, Clavel, and Suchanek}]{helwe2021reasoning}
Chadi Helwe, Chlo{\'e} Clavel, and Fabian~M. Suchanek. 2021.
\newblock Reasoning with transformer-based models: Deep learning, but shallow reasoning.
\newblock In \emph{AKBC}.

\bibitem[{Huang et~al.(2022)Huang, Abbeel, Pathak, and Mordatch}]{huang2022language}
Wenlong Huang, Pieter Abbeel, Deepak Pathak, and Igor Mordatch. 2022.
\newblock Language models as zero-shot planners: Extracting actionable knowledge for embodied agents.
\newblock \emph{arXiv preprint arXiv:2201.07207}.

\bibitem[{Jiang et~al.(2020)Jiang, Xu, Araki, and Neubig}]{jiang2020can}
Zhengbao Jiang, Frank~F Xu, Jun Araki, and Graham Neubig. 2020.
\newblock How can we know what language models know?
\newblock \emph{TACL}.

\bibitem[{Karpukhin et~al.(2020)Karpukhin, O{\u{g}}uz, Min, Lewis, Wu, Edunov, Chen, and Yih}]{karpukhin2020dense}
Vladimir Karpukhin, Barlas O{\u{g}}uz, Sewon Min, Patrick Lewis, Ledell Wu, Sergey Edunov, Danqi Chen, and Wen-tau Yih. 2020.
\newblock Dense passage retrieval for open-domain question answering.
\newblock \emph{arXiv preprint arXiv:2004.04906}.

\bibitem[{Kassner et~al.(2020)Kassner, Krojer, and Sch{\"u}tze}]{kassner2020pretrained}
Nora Kassner, Benno Krojer, and Hinrich Sch{\"u}tze. 2020.
\newblock Are pretrained language models symbolic reasoners over knowledge?
\newblock \emph{arXiv preprint arXiv:2006.10413}.

\bibitem[{Kenton and Toutanova(2019)}]{kenton2019bert}
Jacob Devlin Ming-Wei~Chang Kenton and Lee~Kristina Toutanova. 2019.
\newblock Bert: Pre-training of deep bidirectional transformers for language understanding.
\newblock In \emph{NAACL-HLT}.

\bibitem[{Keysers et~al.(2019)Keysers, Sch{\"a}rli, Scales, Buisman, Furrer, Kashubin, Momchev, Sinopalnikov, Stafiniak, Tihon et~al.}]{keysers2019measuring}
Daniel Keysers, Nathanael Sch{\"a}rli, Nathan Scales, Hylke Buisman, Daniel Furrer, Sergii Kashubin, Nikola Momchev, Danila Sinopalnikov, Lukasz Stafiniak, Tibor Tihon, et~al. 2019.
\newblock Measuring compositional generalization: A comprehensive method on realistic data.
\newblock In \emph{ICLR}.

\bibitem[{Khandelwal et~al.(2019)Khandelwal, Levy, Jurafsky, Zettlemoyer, and Lewis}]{khandelwal2019generalization}
Urvashi Khandelwal, Omer Levy, Dan Jurafsky, Luke Zettlemoyer, and Mike Lewis. 2019.
\newblock Generalization through memorization: Nearest neighbor language models.
\newblock \emph{arXiv preprint arXiv:1911.00172}.

\bibitem[{Kojima et~al.(2022)Kojima, Gu, Reid, Matsuo, and Iwasawa}]{zero2022kojima}
Takeshi Kojima, Shixiang~Shane Gu, Machel Reid, Yutaka Matsuo, and Yusuke Iwasawa. 2022.
\newblock Large language models are zero-shot reasoners.

\bibitem[{Kowalski(1979)}]{kowalski1979algorithm}
Robert Kowalski. 1979.
\newblock Algorithm= logic+ control.
\newblock \emph{Communications of the ACM}.

\bibitem[{Lake and Baroni(2018)}]{lake2018generalization}
Brenden Lake and Marco Baroni. 2018.
\newblock Generalization without systematicity: On the compositional skills of sequence-to-sequence recurrent networks.
\newblock In \emph{ICML}.

\bibitem[{Lewis et~al.(2020)Lewis, Perez, Piktus, Petroni, Karpukhin, Goyal, K{\"u}ttler, Lewis, Yih, Rockt{\"a}schel et~al.}]{lewis2020retrieval}
Patrick Lewis, Ethan Perez, Aleksandra Piktus, Fabio Petroni, Vladimir Karpukhin, Naman Goyal, Heinrich K{\"u}ttler, Mike Lewis, Wen-tau Yih, Tim Rockt{\"a}schel, et~al. 2020.
\newblock Retrieval-augmented generation for knowledge-intensive nlp tasks.
\newblock \emph{Advances in Neural Information Processing Systems}, 33:9459--9474.

\bibitem[{Li et~al.(2022)Li, Puig, Du, Wang, Akyurek, Torralba, Andreas, and Mordatch}]{li2022pretrained}
Shuang Li, Xavier Puig, Yilun Du, Clinton Wang, Ekin Akyurek, Antonio Torralba, Jacob Andreas, and Igor Mordatch. 2022.
\newblock Pre-trained language models for interactive decision-making.
\newblock \emph{arXiv preprint arXiv:2202.01771}.

\bibitem[{Lin et~al.(2015)Lin, Liu, Sun, Liu, and Zhu}]{lin2015learning}
Yankai Lin, Zhiyuan Liu, Maosong Sun, Yang Liu, and Xuan Zhu. 2015.
\newblock Learning entity and relation embeddings for knowledge graph completion.
\newblock In \emph{AAAI}.

\bibitem[{Liu et~al.(2022)Liu, Ash, Goel, Krishnamurthy, and Zhang}]{liu2022transformers}
Bingbin Liu, Jordan~T Ash, Surbhi Goel, Akshay Krishnamurthy, and Cyril Zhang. 2022.
\newblock Transformers learn shortcuts to automata.
\newblock \emph{arXiv preprint arXiv:2210.10749}.

\bibitem[{Liu et~al.(2021)Liu, Shen, Zhang, Dolan, Carin, and Chen}]{liu2021makes}
Jiachang Liu, Dinghan Shen, Yizhe Zhang, Bill Dolan, Lawrence Carin, and Weizhu Chen. 2021.
\newblock What makes good in-context examples for gpt-3?
\newblock \emph{arXiv preprint arXiv:2101.06804}.

\bibitem[{Lu et~al.(2021{\natexlab{a}})Lu, Grover, Abbeel, and Mordatch}]{lu2021pretrained}
Kevin Lu, Aditya Grover, Pieter Abbeel, and Igor Mordatch. 2021{\natexlab{a}}.
\newblock Pretrained transformers as universal computation engines.
\newblock \emph{arXiv preprint arXiv:2103.05247}.

\bibitem[{Lu et~al.(2022)Lu, Welleck, West, Jiang, Kasai, Khashabi, Bras, Qin, Yu, Zellers et~al.}]{lu2021neurologic}
Ximing Lu, Sean Welleck, Peter West, Liwei Jiang, Jungo Kasai, Daniel Khashabi, Ronan~Le Bras, Lianhui Qin, Youngjae Yu, Rowan Zellers, et~al. 2022.
\newblock Neurologic a* esque decoding: Constrained text generation with lookahead heuristics.
\newblock \emph{NAACL}.

\bibitem[{Lu et~al.(2021{\natexlab{b}})Lu, Bartolo, Moore, Riedel, and Stenetorp}]{lu2021fantastically}
Yao Lu, Max Bartolo, Alastair Moore, Sebastian Riedel, and Pontus Stenetorp. 2021{\natexlab{b}}.
\newblock Fantastically ordered prompts and where to find them: Overcoming few-shot prompt order sensitivity.
\newblock \emph{arXiv preprint arXiv:2104.08786}.

\bibitem[{Manning et~al.(2020)Manning, Clark, Hewitt, Khandelwal, and Levy}]{manning2020emergent}
Christopher~D Manning, Kevin Clark, John Hewitt, Urvashi Khandelwal, and Omer Levy. 2020.
\newblock Emergent linguistic structure in artificial neural networks trained by self-supervision.
\newblock \emph{PNAS}.

\bibitem[{Min et~al.(2022)Min, Lyu, Holtzman, Artetxe, Lewis, Hajishirzi, and Zettlemoyer}]{min2022rethinking}
Sewon Min, Xinxi Lyu, Ari Holtzman, Mikel Artetxe, Mike Lewis, Hannaneh Hajishirzi, and Luke Zettlemoyer. 2022.
\newblock Rethinking the role of demonstrations: What makes in-context learning work?
\newblock \emph{arXiv preprint arXiv:2202.12837}.

\bibitem[{Minervini et~al.(2018)Minervini, Bosnjak, Rockt{\"a}schel, and Riedel}]{minervini2018towards}
Pasquale Minervini, Matko Bosnjak, Tim Rockt{\"a}schel, and Sebastian Riedel. 2018.
\newblock Towards neural theorem proving at scale.
\newblock \emph{arXiv preprint arXiv:1807.08204}.

\bibitem[{Minervini et~al.(2020)Minervini, Riedel, Stenetorp, Grefenstette, and Rockt{\"a}schel}]{minervini2020learning}
Pasquale Minervini, Sebastian Riedel, Pontus Stenetorp, Edward Grefenstette, and Tim Rockt{\"a}schel. 2020.
\newblock Learning reasoning strategies in end-to-end differentiable proving.
\newblock In \emph{ICML}.

\bibitem[{Muggleton and De~Raedt(1994)}]{muggleton1994inductive}
Stephen Muggleton and Luc De~Raedt. 1994.
\newblock Inductive logic programming: Theory and methods.
\newblock \emph{The Journal of Logic Programming}.

\bibitem[{Nye et~al.(2021)Nye, Andreassen, Gur-Ari, Michalewski, Austin, Bieber, Dohan, Lewkowycz, Bosma, Luan et~al.}]{nye2021show}
Maxwell Nye, Anders~Johan Andreassen, Guy Gur-Ari, Henryk Michalewski, Jacob Austin, David Bieber, David Dohan, Aitor Lewkowycz, Maarten Bosma, David Luan, et~al. 2021.
\newblock Show your work: Scratchpads for intermediate computation with language models.
\newblock \emph{arXiv preprint arXiv:2112.00114}.

\bibitem[{Patel and Pavlick(2021)}]{patel2021mapping}
Roma Patel and Ellie Pavlick. 2021.
\newblock Mapping language models to grounded conceptual spaces.
\newblock In \emph{ICLR}.

\bibitem[{Petroni et~al.(2019)Petroni, Rockt{\"a}schel, Riedel, Lewis, Bakhtin, Wu, and Miller}]{petroni2019language}
Fabio Petroni, Tim Rockt{\"a}schel, Sebastian Riedel, Patrick Lewis, Anton Bakhtin, Yuxiang Wu, and Alexander Miller. 2019.
\newblock Language models as knowledge bases?
\newblock In \emph{EMNLP-IJCNLP}.

\bibitem[{Polu and Sutskever(2020)}]{polu2020generative}
Stanislas Polu and Ilya Sutskever. 2020.
\newblock Generative language modeling for automated theorem proving.
\newblock \emph{arXiv preprint arXiv:2009.03393}.

\bibitem[{Press et~al.(2021)Press, Smith, and Lewis}]{press2021train}
Ofir Press, Noah~A Smith, and Mike Lewis. 2021.
\newblock Train short, test long: Attention with linear biases enables input length extrapolation.
\newblock \emph{arXiv preprint arXiv:2108.12409}.

\bibitem[{Press et~al.(2022)Press, Zhang, Min, Schmidt, Smith, and Lewis}]{press2022measuring}
Ofir Press, Muru Zhang, Sewon Min, Ludwig Schmidt, Noah~A Smith, and Mike Lewis. 2022.
\newblock Measuring and narrowing the compositionality gap in language models.
\newblock \emph{arXiv preprint arXiv:2210.03350}.

\bibitem[{Qu et~al.(2020)Qu, Chen, Xhonneux, Bengio, and Tang}]{qu2020rnnlogic}
Meng Qu, Junkun Chen, Louis-Pascal Xhonneux, Yoshua Bengio, and Jian Tang. 2020.
\newblock Rnnlogic: Learning logic rules for reasoning on knowledge graphs.
\newblock In \emph{ICLR}.

\bibitem[{Radford et~al.(2019)Radford, Wu, Child, Luan, Amodei, Sutskever et~al.}]{radford2019language}
Alec Radford, Jeffrey Wu, Rewon Child, David Luan, Dario Amodei, Ilya Sutskever, et~al. 2019.
\newblock Language models are unsupervised multitask learners.
\newblock \emph{OpenAI blog}.

\bibitem[{Razeghi et~al.(2022)Razeghi, Logan~IV, Gardner, and Singh}]{razeghi2022impact}
Yasaman Razeghi, Robert~L Logan~IV, Matt Gardner, and Sameer Singh. 2022.
\newblock Impact of pretraining term frequencies on few-shot reasoning.
\newblock \emph{arXiv preprint arXiv:2202.07206}.

\bibitem[{Reimers and Gurevych(2019)}]{reimers2019sentence}
Nils Reimers and Iryna Gurevych. 2019.
\newblock Sentence-bert: Sentence embeddings using siamese bert-networks.
\newblock In \emph{EMNLP-IJCNLP}.

\bibitem[{Roberts et~al.(2020)Roberts, Raffel, and Shazeer}]{roberts2020much}
Adam Roberts, Colin Raffel, and Noam Shazeer. 2020.
\newblock How much knowledge can you pack into the parameters of a language model?
\newblock In \emph{EMNLP}.

\bibitem[{Rockt{\"a}schel and Riedel(2017)}]{rocktaschel2017end}
Tim Rockt{\"a}schel and Sebastian Riedel. 2017.
\newblock End-to-end differentiable proving.
\newblock \emph{NeurIPS}.

\bibitem[{Saparov and He(2022)}]{saparov2022language}
Abulhair Saparov and He~He. 2022.
\newblock Language models are greedy reasoners: A systematic formal analysis of chain-of-thought.
\newblock \emph{arXiv preprint arXiv:2210.01240}.

\bibitem[{Shao et~al.(2023)Shao, Gong, Shen, Huang, Duan, and Chen}]{shao2023enhancing}
Zhihong Shao, Yeyun Gong, Yelong Shen, Minlie Huang, Nan Duan, and Weizhu Chen. 2023.
\newblock Enhancing retrieval-augmented large language models with iterative retrieval-generation synergy.
\newblock \emph{arXiv preprint arXiv:2305.15294}.

\bibitem[{Shin et~al.(2022)Shin, Lee, Ahn, Kim, Kim, Kim, Cho, Lee, Park, Ha et~al.}]{shin2022effect}
Seongjin Shin, Sang-Woo Lee, Hwijeen Ahn, Sungdong Kim, HyoungSeok Kim, Boseop Kim, Kyunghyun Cho, Gichang Lee, Woomyoung Park, Jung-Woo Ha, et~al. 2022.
\newblock On the effect of pretraining corpora on in-context learning by a large-scale language model.
\newblock \emph{arXiv preprint arXiv:2204.13509}.

\bibitem[{Shuster et~al.(2021)Shuster, Poff, Chen, Kiela, and Weston}]{shuster2021retrieval}
Kurt Shuster, Spencer Poff, Moya Chen, Douwe Kiela, and Jason Weston. 2021.
\newblock Retrieval augmentation reduces hallucination in conversation.
\newblock \emph{arXiv preprint arXiv:2104.07567}.

\bibitem[{Sinha et~al.(2019)Sinha, Sodhani, Dong, Pineau, and Hamilton}]{sinha2019clutrr}
Koustuv Sinha, Shagun Sodhani, Jin Dong, Joelle Pineau, and William~L. Hamilton. 2019.
\newblock Clutrr: A diagnostic benchmark for inductive reasoning from text.
\newblock \emph{EMNLP}.

\bibitem[{Talmor et~al.(2020)Talmor, Elazar, Goldberg, and Berant}]{talmor2020olmpics}
Alon Talmor, Yanai Elazar, Yoav Goldberg, and Jonathan Berant. 2020.
\newblock olmpics-on what language model pre-training captures.
\newblock \emph{TACL}.

\bibitem[{Touvron et~al.(2023)Touvron, Martin, Stone, Albert, Almahairi, Babaei, Bashlykov, Batra, Bhargava, Bhosale et~al.}]{touvron2023llama}
Hugo Touvron, Louis Martin, Kevin Stone, Peter Albert, Amjad Almahairi, Yasmine Babaei, Nikolay Bashlykov, Soumya Batra, Prajjwal Bhargava, Shruti Bhosale, et~al. 2023.
\newblock Llama 2: Open foundation and fine-tuned chat models.
\newblock \emph{arXiv preprint arXiv:2307.09288}.

\bibitem[{Wei et~al.(2022{\natexlab{a}})Wei, Tay, Bommasani, Raffel, Zoph, Borgeaud, Yogatama, Bosma, Zhou, Metzler et~al.}]{wei2022emergent}
Jason Wei, Yi~Tay, Rishi Bommasani, Colin Raffel, Barret Zoph, Sebastian Borgeaud, Dani Yogatama, Maarten Bosma, Denny Zhou, Donald Metzler, et~al. 2022{\natexlab{a}}.
\newblock Emergent abilities of large language models.
\newblock \emph{TMLR}.

\bibitem[{Wei et~al.(2022{\natexlab{b}})Wei, Wang, Schuurmans, Bosma, Chi, Le, and Zhou}]{wei2022chain}
Jason Wei, Xuezhi Wang, Dale Schuurmans, Maarten Bosma, Ed~Chi, Quoc Le, and Denny Zhou. 2022{\natexlab{b}}.
\newblock Chain of thought prompting elicits reasoning in large language models.
\newblock \emph{arXiv preprint arXiv:2201.11903}.

\bibitem[{Wolf et~al.(2019)Wolf, Debut, Sanh, Chaumond, Delangue, Moi, Cistac, Rault, Louf, Funtowicz et~al.}]{wolf2019huggingface}
Thomas Wolf, Lysandre Debut, Victor Sanh, Julien Chaumond, Clement Delangue, Anthony Moi, Pierric Cistac, Tim Rault, R{\'e}mi Louf, Morgan Funtowicz, et~al. 2019.
\newblock Huggingface's transformers: State-of-the-art natural language processing.
\newblock \emph{arXiv preprint arXiv:1910.03771}.

\bibitem[{Wu et~al.(2021)Wu, Rabe, Li, Ba, Grosse, and Szegedy}]{wu2021lime}
Yuhuai Wu, Markus~N Rabe, Wenda Li, Jimmy Ba, Roger~B Grosse, and Christian Szegedy. 2021.
\newblock Lime: Learning inductive bias for primitives of mathematical reasoning.
\newblock In \emph{ICML}.

\bibitem[{Xie et~al.(2021)Xie, Raghunathan, Liang, and Ma}]{xie2021explanation}
Sang~Michael Xie, Aditi Raghunathan, Percy Liang, and Tengyu Ma. 2021.
\newblock An explanation of in-context learning as implicit bayesian inference.
\newblock \emph{ICLR}.

\bibitem[{Yang et~al.(2017)Yang, Yang, and Cohen}]{yang2017differentiable}
Fan Yang, Zhilin Yang, and William~W Cohen. 2017.
\newblock Differentiable learning of logical rules for knowledge base reasoning.
\newblock \emph{NeurIPS}.

\bibitem[{Yang and Song(2020)}]{yang2020learn}
Yuan Yang and Le~Song. 2020.
\newblock Learn to explain efficiently via neural logic inductive learning.
\newblock In \emph{ICLR}.

\bibitem[{Zelikman et~al.(2022)Zelikman, Wu, and Goodman}]{zelikman2022star}
Eric Zelikman, Yuhuai Wu, and Noah~D Goodman. 2022.
\newblock Star: Bootstrapping reasoning with reasoning.
\newblock \emph{arXiv preprint arXiv:2203.14465}.

\bibitem[{Zhang et~al.(2022)Zhang, Li, Huang, Naik, and Xing}]{zhang2022improved}
Hanlin Zhang, Ziyang Li, Jiani Huang, Mayur Naik, and Eric Xing. 2022.
\newblock Improved logical reasoning of language models via differentiable symbolic programming.
\newblock In \emph{First Workshop on Pre-training: Perspectives, Pitfalls, and Paths Forward at ICML 2022}.

\bibitem[{Zhao et~al.(2021)Zhao, Wallace, Feng, Klein, and Singh}]{zhao2021calibrate}
Zihao Zhao, Eric Wallace, Shi Feng, Dan Klein, and Sameer Singh. 2021.
\newblock Calibrate before use: Improving few-shot performance of language models.
\newblock In \emph{ICML}.

\bibitem[{Zhou et~al.(2022{\natexlab{a}})Zhou, Sch{\"a}rli, Hou, Wei, Scales, Wang, Schuurmans, Bousquet, Le, and Chi}]{zhou2022least}
Denny Zhou, Nathanael Sch{\"a}rli, Le~Hou, Jason Wei, Nathan Scales, Xuezhi Wang, Dale Schuurmans, Olivier Bousquet, Quoc Le, and Ed~Chi. 2022{\natexlab{a}}.
\newblock Least-to-most prompting enables complex reasoning in large language models.
\newblock \emph{arXiv preprint arXiv:2205.10625}.

\bibitem[{Zhou et~al.(2022{\natexlab{b}})Zhou, Nova, Larochelle, Courville, Neyshabur, and Sedghi}]{zhou2022teaching}
Hattie Zhou, Azade Nova, Hugo Larochelle, Aaron Courville, Behnam Neyshabur, and Hanie Sedghi. 2022{\natexlab{b}}.
\newblock Teaching algorithmic reasoning via in-context learning.
\newblock \emph{arXiv preprint arXiv:2211.09066}.

\bibitem[{Zhou et~al.(2020)Zhou, Hu, Zhang, Liang, Sun, Xiong, and Tang}]{zhou2020towards}
Wangchunshu Zhou, Jinyi Hu, Hanlin Zhang, Xiaodan Liang, Maosong Sun, Chenyan Xiong, and Jian Tang. 2020.
\newblock Towards interpretable natural language understanding with explanations as latent variables.
\newblock \emph{NeurIPS}.

\end{thebibliography}
\bibliographystyle{acl2023/acl_natbib}

\clearpage
\appendix
{\LARGE \textbf{Appendix}}
\section{Extended Related Work}
\label{app:related}
\textbf{Neuro-Symbolic Reasoning.} ILP \citep{muggleton1994inductive} and its neural version \citep{yang2020learn} are unable to reason about disjoint relations in confront of missing links when KBs are noisy like in FreeBase, which means ILP only synthesizes rules based on existing relations. 
Methods like Neural-LP \citep{yang2017differentiable} and RNNLogic \citep{qu2020rnnlogic} require enumeration of all possible rules given a max rule length T. Thus the complexity of these models grows exponentially as maximum rule length increases, which is a significant disadvantage for systematicity problems. 
For deductive reasoning, NTP \citep{rocktaschel2017end} and its improved versions \citep{minervini2018towards, minervini2020learning} require hand-crafted templates to imitate backward chaining for deductive reasoning. 
This belies the considerable user burden of authoring the templates which then fundamentally biases the tool toward a specific subset of programs that the author has in mind. 
Moreover, the performance and efficiency of NTP is far from satisfactory: the performance usually lags far behind its neural counterparts like knowledge graph embedding methods \citep{lin2015learning}; during both training and inference, NTPs need to compute all possible proof trees needed for proving a query, relying on the continuous unification of the query with all the rules and facts in the KB. The search space of existing works is exponentially large, which makes them hard to scale up in general \citep{minervini2018towards, chaudhuri2021neurosymbolic}.

\textbf{LMs for Theorem Proving.} Most works focus on proving formal mathematical theorems: 
GPT-f \citep{polu2020generative} shows promising results by generative language modeling over mathematical formulas. Systematicity of LMs when training on proofs is evaluated in \citep{gontier2020measuring} but shows negative results in generalizing to unseen proof steps in extrapolation and complex language. 
Three synthetic tasks inspired by three reasoning primitives of deduction, induction, and abduction are demonstrated in \citep{wu2021lime}. The above works provide insights into understanding LMs' reasoning capabilities. 
Though they share similar problem structures like compositionality with ours, they fundamentally require large-scale pre-training and fine-tuning due to the mismatch between Wikipedia pre-training corpora and mathematical formulas. Such a re-training requirement not only results in computational inefficiency but lacking in compositional generalization to longer proof steps unseen during training \citep{gontier2020measuring}.

\textbf{Symbolic Reasoning with LMs.} 
Large LMs pre-trained on open-domain text corpora have achieved impressive advances in natural language generation and understanding tasks \citep{kenton2019bert, brown2020language}. 
By self-supervised imitation on human-generated texts, LMs contain rich factual knowledge \citep{petroni2019language, bouraoui2020inducing, roberts2020much} and linguistic structures \citep{manning2020emergent}, serving as a versatile inference regime for various downstream tasks \citep{brown2020language, lu2021pretrained}. 
Among them, GPT-3 stands out by its few-shot generalization to unseen cases without further fine-tuning given in-context samples as demonstrations \citep{brown2020language}. 
Constraint decoding is shown to be effective in incorporating logical constraints into natural language generation \citep{lu2021neurologic}.
However, it is a common belief that LMs have not yet enjoyed a comparable success in tasks that require extensive planning and grounding \citep{glenberg2002grounding, bender2020climbing, bisk2020experience} as well as symbolic reasoning \citep{kassner2020pretrained, helwe2021reasoning, razeghi2022impact}.

\section{Algorithm Description}
\label{app:algo}
Algorithm~\ref{algo:main} describes the procedure or \abbr. It can also be illustrated in Figure~\ref{fig:model}.

\begin{algorithm*}
\caption{Generate proof path from Pre-Trained Language Models.}
\begin{algorithmic}
\Require {Planning LM $\mathcal{P}_\theta$, Translation LM $\mathcal{T}_\phi$, Query set $\mathcal{Q}$ that contains all query triplets, $\F$ that contains all available facts, $\R$ that contains all the available logic rules or proof examples.}
\For{$q=(s,p,o)\in\mathcal{Q}$} // \textit{$s,p,o$ denote subject entity, predicate (relation) and object entity respectively.}
\State Find $r\in\R$, whose task relation is $p$.
\State Construct prompt $r'=[r,q]$.~\;\;// \textit{$[r,q]$ means the concatenation of two strings.}
\While{Max step is not reached}
\State Sample $10$ sentences $\{x_i\}_{i=1}^{10}$ from $\mathcal{P}_\theta(f')$.
\State Set $\F'\in\F$ whose first entity are $s$.
\If{$|\F'|==0$}
\State Break ~\;\;// \textit{No available facts in the $\F$ start with entity $s$.}
\EndIf
\For{$x\in\{x_i\}_{i=1}^{10}$}
\State $score_i = \max_{\forall r\in\F'} cosine(\mathcal{T}_\phi(x),\mathcal{T}_\phi(r));$// \textit{Cosine similarities of $s$ to facts in $\F'$.}
\State $idx=\arg\max_{\forall r\in\F'} cosine(\mathcal{T}_\phi(x),\mathcal{T}_\phi(r));$// \textit{Select $r\in\F'$ with the highest similarity to $x$.}
\State $x'=\F'[idx]$ 
\EndFor
\State Choose the highest score rule $x^*$ as the next proof step and append it to the prompt $f'=[f',x^*]$.
\If{$o'==o$}
\State Break \;\;// \textit{The object entity converges to the target entity $o$.}
\EndIf
\EndWhile
\EndFor
\end{algorithmic}
\label{algo:main}
\end{algorithm*}

\subsection{Data Generation.}\label{app:data}

\textbf{\clutrr.} 
CLUTRR has 9 subsets with difference story length, named $l_2,l_3, \dots, l_{10}$. Following~\citep{minervini2020learning}, we convert $l_2,l_3,l_4$ to the \textbf{$\R$} and use $l_5,\dots,l_{10}$ to the \textbf{query sets}. As illustrated in Table.~\ref{tab:data_process}, data samples in CLUTRR consist of a story and a target, where the target contains two entities and the relation that is needed to be inferred, the story contains available triplets. Each sample in the $l_2,l_3,l_4$ will be converted to the format ``Task: \dots, Step i: \dots'' and added to the $\R$. Note that all examples in the $\R$ have a story length of less than five, which enables us to test the systematic generalization ability of \abbr. For CLUTRR, the story triplets in the $\R$ are not useful for test target proving, because they are all from different relation graphs. For example, story triplets in the $l_2,l_3,l_4$ contain ``(William's brother is Steve)'' while one test story on $l_5$ contains ``(William's uncle is Steve)''. During the evaluation, if the model chooses ``(William's brother is Steve)'', the proof path will be wrong. However, the similarity of these two triplets is high, the model is then easy to make errors and these noisy facts increase proof difficulties. We hence evaluate our methods in two settings considering the number of noisy facts. The simplest setting (\textbf{Test Facts Setting}) is that, when queries are from $l_i,i\in[5,\dots,10]$, the $\F$ only contains facts in $l_i$. In this case, the $\F_{5\sim10}$ have 251,222,275,279,285,304 facts respectively. 
The most difficult setting is termed \textbf{All Facts Setting}. We first extract facts in the $\F$ with length $l_2,l_3,l_4$ and get totally $5,210$ facts. When queries are from $l_i,i\in[5,\dots,10]$, the $\F$ contains triplets in $l_i,l_2,l_3,l_4$, where the additional $5,210$ facts are not useful for the proof path and are noisy facts. The All Facts Setting is set as our default setting and experimental results of the Test Facts Setting are mainly in the Appendix. 
For CoT, the $\F$ is needless and the construction of prompt examples is slightly different from the procedure above. Specifically, as shown in~\figurename~\ref{fig:model_cot}, for each target in the training samples, we need to preserve the story and extract a proof path for the target. 

\textbf{\countries.} 
Training samples in Countries are triplets that describe the \textit{neighbor of} relation or \textit{located in} relation of two regions/subregions/countries and can thus be directly used as \textbf{$\F$}.  Because the three tasks (\textit{S1, S2, S3}) ~\citep{minervini2020learning} have different training sets and thus have different $\F$. Test samples in Countries are also triplets with specific entities and relations, hence the \textbf{query set} is just the test set of the original Countries dataset. One main difficulty in applying the proposed method to Countries is the lack of off-the-shelf proof paths ($\R$). The CTP~\citep{minervini2020learning} model is trained and used for proving each triplet in the training set. CTP returns the scores of the possible proof path and the proof with the maximum score is iteratively searched as added into the \textbf{$\R$}. After that, $924, 906, 705$ available examples are found for \textit{S1, S2, S3} tasks respectively.

\section{Additional Experimental Setups and Results}
\label{app:exp}

\begin{table*}[h]
\centering
\adjustbox{max width=0.85\textwidth}{%
\begin{tabular}{l|l}
 \toprule
\multicolumn{1}{c}{\textbf{CoT Prompting}} &            \multicolumn{1}{c}{\textbf{\abbr}}                                                  \\\midrule
\begin{tabular}[c]{@{}l@{}}\textcolor{cyan}{Question:} Wilhelmina took her uncle Hugh to the grocery store. Francisco and \\his brother Wesley were wrestling. Wilhelmina, Francisco's daughter, was \\cheering on the competition. What is the relation between Hugh and Wesley?\\
\textcolor{cyan}{Answer:} \textcolor{orange}{Wesley's brother is Francisco, Francisco's daughter is Wilhelmina,}\\ \textcolor{orange}{ Wilhelmina's uncle is Hugh.} The relation of Hugh between Wesley is brother.\\  \\ 
\textcolor{cyan}{Question:} Constance went shoe shopping with her sister Ellen. Elsie had a \\ daughter named Constance. Elsie had picked her daughter Margaret out \\ the cutest new dress to wear on her birthday. Charles and his sister Kathleen \\ have been best friends ever since childhood. Nadia and her father, James, \\ went to the marina. James's daughter, Mabel, had purchased a boat, and \\ they were eager to see it. Mabel bought her mother, Ellen, a puppy for her \\ birthday. James hung his son Charles's finger paintings on the refrigerator. \\ The paintings were right next to the paintings of Nadia, Charles's sister. \\ Kathleen wasn't old enough to make any paintings for her father, James. \\ What is the relation between Margaret and Charles?\\ \textcolor{cyan}{Answer:} \textcolor{orange}{charles' mother is nancy, nancy's daughter is elizabeth,}\\ \textcolor{orange}{elizabeth's husband is john, john's wife is mary, mary's brother is george,}\\ \textcolor{orange}{david's nephews are william, william's nephews are robert.}\wrong\end{tabular}                          & \makecell[l]{\Task: A's aunt is B
 \\
\textcolor{orange}{Step 1}: A's sister is C \\
\textcolor{orange}{Step 2}: C's mother is D \\
\textcolor{orange}{Step 3}: D's sister is B \\
\Task: Carrie's aunt is Lynn \\ 
\textcolor{orange}{Step 1}: Carrie's sister is Natasha \\
\textcolor{orange}{Step 2}: Natasha's mother is Cindy \\
\textcolor{orange}{Step 3}: Cindy's sister is Lynn \\\\
\Task: Charles's aunt is Margaret \\
\textcolor{orange}{Step 1}: Charles's sister is Nadia \\
\textcolor{orange}{Step 2}: Nadia's aunt is Margaret  \coloright
} \\ \bottomrule
                
\end{tabular}}
\caption{Qualitative comparison of CoT and \abbr~over the same example on \clutrr.}
\label{tab:lm_cot}
\end{table*}

\textbf{Human evaluation results on the \countries~benchmark.} Because \countries is a simple reasoning benchmark, which has only two predicates, namely \sclcode{LocatedIN}, and \sclcode{NeighborOf}. Table.~\ref{tab:human_countries} shows that simple baselines with entity-based prompts or no prompt can attain comparable or higher results than \abbr on the simplest task $S1$. However, \abbr attains better reasoning capability on harder tasks. Besides, All baselines outperform COT by a large margin, which verifies the importance of generating a reasoning path step by step than proofing the query predicate at once.

\begin{table*}[h]
\centering
\adjustbox{max width=.95\textwidth}{%
\begin{tabular}{c|c|c}
\toprule
     \textbf{Method} & \textbf{Prompt} & \textbf{Output} \\ \hline
Facts w/o Provenance %
& \makecell[l]{ 
\Task: Kazakhstan locatedIn Asia }
& \makecell[l]{
\textcolor{orange}{Step 1}: Kazakhstan locatedIn CentralAsia \\
\textcolor{orange}{Step 2}: CentralAsia locatedIn Asia } \\ \hline

Facts w/ Provenance
& \makecell[l]{\Task: Kazakhstan locatedIn CentralAsia \\
\textcolor{orange}{Step 1}: Kazakhstan neighborOf Uzbekistan \\
\textcolor{orange}{Step 2}: Uzbekistan locatedIn CentralAsia \\
\Task: Kazakhstan locatedIn Asia} 
& \makecell[l]{
\textcolor{orange}{Step 1}: Kazakhstan neighborOf China \\
\textcolor{orange}{Step 2}: China neighborOf India \\
\textcolor{orange}{Step 3}: India locatedIn SouthernAsia \\
\textcolor{orange}{Step 4}: SouthernAsia locatedIn Asia } \\ \hline

\multirow{18}{*}{\abbr~(N=1)} %
& \makecell[l]{ 
\Task: X is located in Z \\
\textcolor{orange}{Step 1}: X is located in Y \\
\textcolor{orange}{Step 2}: Y is located in M \\
\textcolor{orange}{Step 3}: M is located in Z \\
\Task: Kazakhstan locatedIn Asia }
& \makecell[l]{
\textcolor{orange}{Step 1}: Kazakhstan neighborOf kyrgyzstan \\
\textcolor{orange}{Step 2}: Kyrgyzstan neighborOf Uzbekistan \\
\textcolor{orange}{Step 3}: Uzbekistan neighborOf Afghanistan \\
\textcolor{orange}{Step 4}: Afghanistan neighborOf china \\
\textcolor{orange}{Step 5}: China neighborOf India \\
\textcolor{orange}{Step 6}: India neighborOf Nepal \\
\textcolor{orange}{Step 7}: Nepal locatedIn Asia } \\ \cline{2-3}

& \makecell[l]{ 
\Task: X locatedIn Y \\
\textcolor{orange}{Step 1}: X neighborOf M \\
\textcolor{orange}{Step 2}: M neighborOf N \\
\textcolor{orange}{Step 3}: N locatedIn Y \\
\Task: Kazakhstan locatedIn Asia}
& \makecell[l]{
\textcolor{orange}{Step 1}: Kazakhstan locatedIn CentralAsia \\
\textcolor{orange}{Step 2}: CentralAsia locatedIn Asia } \\ \cline{2-3}

& \makecell[l]{ 
\Task: X locatedIn Y \\
\textcolor{orange}{Step 1}: X neighborOf M \\
\textcolor{orange}{Step 2}: M locatedIn N \\
\textcolor{orange}{Step 3}: N locatedIn Y \\
\Task: Kazakhstan locatedIn Asia}
& \makecell[l]{
\textcolor{orange}{Step 1}: Kazakhstan locatedIn CentralAsia \\
\textcolor{orange}{Step 2}: CentralAsia locatedIn Asia } \\ \cline{2-3}

& \makecell[l]{ 
\Task: X locatedIn Y \\
\textcolor{orange}{Step 1}: X neighborOf Z \\
\textcolor{orange}{Step 2}: Z locatedIn Y \\
\Task: Kazakhstan locatedIn Asia }
& \makecell[l]{
\textcolor{orange}{Step 1}: Kazakhstan neighborOf kyrgyzstan\\
\textcolor{orange}{Step 2}: Kyrgyzstan neighborOf Uzbekistan \\
\textcolor{orange}{Step 3}: Uzbekistan neighborOf Afghanistan \\
\textcolor{orange}{Step 4}: Afghanistan neighborOf Tajikistan \\
\textcolor{orange}{Step 5}: Tajikistan neighborOf China \\
\textcolor{orange}{Step 6}: China neighborOf India \\
\textcolor{orange}{Step 7}: India neighborOf Nepal \\
\textcolor{orange}{Step 8}: Nepal locatedIn Asia } %

\\ \hline
\end{tabular}}
\caption{Sensitivity analysis of prompt formats using \abbr~in the Countries dataset \citep{rocktaschel2017end}.}
\label{tab:sensitivity}
\end{table*}

\begin{table*}[h]
\centering
\adjustbox{max width=.95\textwidth}{%
\begin{tabular}{c|c|c}
\toprule
     \textbf{Method} & \textbf{Prompt} & \textbf{Output} \\ \hline

No Prompt 
& \makecell[l]{ %
\Task: Richard's father is Donald}
& \makecell[l]{
\textcolor{orange}{Step 1}: Richard's grandmother is Elizabeth \\
\textcolor{orange}{Step 2}: Elizabeth's son is Donald } \\ \hline

Random Prompt 
&  \makecell[l]{ %
\Task: C's father is B \\
\textcolor{orange}{Step 1}: C's brother is A \\
\textcolor{orange}{Step 2}: A's father is B \\
\Task: Janet's father is George \\
\textcolor{orange}{Step 1}: Janet's brother is John \\
\textcolor{orange}{Step 2}: John's father is George \\ 
\Task: Pat's grandfather is Jose }
& \makecell[l]{ %
\textcolor{orange}{Step 1}: Pat's uncle is Samuel \\ 
\textcolor{orange}{Step 2}: Samuel's father is Jose } \\ \hline

Rule Only  %
& \makecell[l]{ \Task: B's father is C \\
\textcolor{orange}{Step 1}: B's mother is D \\
\textcolor{orange}{Step 2}: D's son is A \\
\textcolor{orange}{Step 3}: A's father is C \\
\Task: Richard's father is Donald } 
& \makecell[l]{ %
\textcolor{orange}{Step 1}: Richard's grandmother is Elizabeth \\
\textcolor{orange}{Step 2}: Elizabeth's son is Donald }
\\ \hline

Entity-based Prompt  %
& \makecell[l]{ \Task: C's brother is A \\
\textcolor{orange}{Step 1}: C's brother is B \\
\textcolor{orange}{Step 2}: B's brother is A \\
\Task: Milton's brother is Glen \\
\textcolor{orange}{Step 1}: Milton's brother is Michael \\
\textcolor{orange}{Step 2}: Michael's brother is Glen \\
\Task: Milton's mother is Antonia }
& \makecell[l]{  %
\textcolor{orange}{Step 1}: Milton's daughter is Margaretta \\
\textcolor{orange}{Step 2}: Margaretta's grandmother is Antonia } 
\\ \hline

Language Planner  
& \makecell[l]{ \Task: David's nephew is Don \wrong \\
\textcolor{orange}{Step 1}: David's mother is Frances \\
\textcolor{orange}{Step 2}: Frances's daughter is Lynn \\
\textcolor{orange}{Step 3}: Lynn's son is Don \\
\Task: Patricia's uncle is Donald }
& \makecell[l]{\textcolor{orange}{Step 1}: Richard's father is Donald } 
\\ \hline

\abbr~(N=1)
& \makecell[l]{\Task: B's father is C \\
\textcolor{orange}{Step 1}: B's mother is D \\
\textcolor{orange}{Step 2}: D's daughter is A \\
\textcolor{orange}{Step 3}: A's father is C \\
\Task: Bobby's father is Hugh \\ 
\textcolor{orange}{Step 1}: Bobby's mother is David \\
\textcolor{orange}{Step 2}: David's daughter is Marie \\
\textcolor{orange}{Step 3}: Marie's father is Hugh \\
\Task: Richard's father is Donald \\
}
& \makecell[l]{
\textcolor{orange}{Step 1}: Richard's mother is Maria \\
\textcolor{orange}{Step 2}: Maria's husband is Donald}
\\ \hline
\abbr~(N=2) %
& \makecell[l]{ \Task: E's father is B \\
\textcolor{orange}{Step 1}: E's daughter is A \\
\textcolor{orange}{Step 2}: A's sister is D \\
\textcolor{orange}{Step 3}: D's brother is C \\
\textcolor{orange}{Step 4}: C's grandfather is B \\
\Task: Louis's father is Harold \\
\textcolor{orange}{Step 1}: Louis's daughter is Jennifer \\
\textcolor{orange}{Step 2}: Jennifer's sister is Nettie \\
\textcolor{orange}{Step 3}: Nettie's brother is Samuel \\
\textcolor{orange}{Step 4}: Samuel's grandfather is Harold \\ 
\Task: D's father is C \\
\textcolor{orange}{Step 1}: D's husband is A \\
\textcolor{orange}{Step 2}: A's son is B \\
\textcolor{orange}{Step 3}: B's grandfather is C \\
\Task: Lena's father is James \\
\textcolor{orange}{Step 1}: Lena's husband is Wesley \\
\textcolor{orange}{Step 2}: Wesley's son is Theodore \\
\textcolor{orange}{Step 3}: Theodore's grandfather is James \\
\Task: Richard's father is Donald }
& \makecell[l]{
\textcolor{orange}{Step 1}: Richard's grandmother is Elizabeth \\
\textcolor{orange}{Step 2}: Elizabeth's son is Donald} 
\\ \hline
\end{tabular}}
\caption{Failure cases analysis - the results of different prompts for the same task.}
\label{tab:comp_prompt}
\end{table*}

\begin{table*}[h]
\adjustbox{max width=.95\textwidth}{%
\begin{tabular}{@{}lll|l@{}}
\toprule
\multicolumn{3}{c}{\textbf{Wrong}} & \multicolumn{1}{c}{\textbf{Right}} \\ \midrule
\begin{tabular}[c]{@{}l@{}}\Task: C's nephew is A\\  
\textcolor{orange}{Step 1}: C's son is D\\  \textcolor{orange}{Step 2}: D's aunt is B\\ \textcolor{orange}{Step 3}: B's son is A\\ \\ \Task: Lena's nephew is Richard\\  \textcolor{orange}{Step 1}: Lena's son is Theodore\\  \textcolor{orange}{Step 2}: Theodore's aunt is Cindy\\  \textcolor{orange}{Step 3}: Cindy's son is Richard\\ \\ \Task: Donald's nephew is Gilbert\\ \textcolor{orange}{Step 1}: Donald's son is Edgar\\
\textcolor{orange}{Step 2}: Edgar's mother is Marge \wrong \\
\textcolor{orange}{Step 3}: Marge's son is
John\end{tabular} & 
\begin{tabular}[c]{@{}l@{}}
\Task: A's nephew is B\\  
\textcolor{orange}{Step 1}: A's brother is C\\ 
\textcolor{orange}{Step 2}: C's son is B\\ \\ 
\Task: Theresa's nephew is John\\  \textcolor{orange}{Step 1}: Theresa's brother is George\\
\textcolor{orange}{Step 2}: George's son is John\\ \\
\Task: Donald's nephew is Gilbert\\
\textcolor{orange}{Step 1}: Donald's brother is James\\
\textcolor{orange}{Step 2}: James's grandson is Johnson\\
\textcolor{orange}{Step 3}: Johnson's father is Hugh\\
\textcolor{orange}{Step 4}: Hugh's brother is Wesley \wrong
\end{tabular} & 
\begin{tabular}[c]{@{}l@{}}\Task: C's nephew is A\\ 
\textcolor{orange}{Step 1}: C's son is D\\ 
\textcolor{orange}{Step 2}: D's aunt is B\\ 
\textcolor{orange}{Step 3}: B's son is A\\ \\ 
\Task:
Lena's nephew is Richard\\  
\textcolor{orange}{Step 1}:
Lena's son is Theodore\\  
\textcolor{orange}{Step 2}:
Theodore's aunt is Cindy\\  
\textcolor{orange}{Step 3}:
Cindy's son is Richard\\ \\ 
\Task: Donald's nephew is Gilbert\\ 
\textcolor{orange}{Step 1}: Donald's son is
Edgar\\
\textcolor{orange}{Step 2}: Edgar's mother is
Marge \wrong 
\end{tabular} & 
\begin{tabular}[c]{@{}l@{}} 
\Task: A's nephew is E \\ 
\textcolor{orange}{Step 1}: A's son is D \\  \textcolor{orange}{Step 2}: D's grandmother is B \\ 
\textcolor{orange}{Step 3}: B's son is C \\ 
\textcolor{orange}{Step 4}: C's son is E \\ \\ 
\Task: Antonio's nephew is Allan\\  
\textcolor{orange}{Step 1}: Antonio's son is Bobby\\ 
\textcolor{orange}{Step 2}: Bobby's grandmother is
Shantel\\  
\textcolor{orange}{Step 3}: Shantel's son is
Robert\\  
\textcolor{orange}{Step 4}: Robert's son is
Allan\\ \\ 
\Task: Donald's nephew is Gilbert\\ 
\textcolor{orange}{Step 1}: Donald's son is Jeremy\\
\textcolor{orange}{Step 2}: Jeremy's grandfather is
Jose\\ 
\textcolor{orange}{Step 3}: Jose's son is Samuel \\ \textcolor{orange}{Step 4}: Samuel's son is Gilbert
\end{tabular} 
\\ \bottomrule
\end{tabular}}
\caption{The comparison of different demonstrations on results.}
\label{tab:example_multi_prompt}
\end{table*}

\begin{table*}[h]
\centering
\adjustbox{max width=.95\textwidth}{%
\begin{tabular}{c|c|c}
\toprule
\textbf{Method} & \textbf{Example 1} & \textbf{Example 2} \\ \hline
Random Prompt &  \makecell[l]{ %
\Task: D's mother is B \\
\textcolor{orange}{Step 1}: D's father is E \\
\textcolor{orange}{Step 2}: E's daughter is C \\
\textcolor{orange}{Step 3}: C's brother is A \\
\textcolor{orange}{Step 4}: A's mother is B \\
\Task: Allan's mother is Marie \\
\textcolor{orange}{Step 1}: Allan's father is Robert \\
\textcolor{orange}{Step 2}: Robert's daughter is Michelle \\ 
\textcolor{orange}{Step 3}: Michelle's brother is Ronald\\
\textcolor{orange}{Step 4}: Ronald's mother is Marie \\
\Task: Jose's granddaughter is Margaretta \\
\textcolor{orange}{Step 1}: Jose's father is William \wrong \\ 
\textcolor{orange}{Step 2}: William's niece is Margaretta
}
& \makecell[l]{ %
\Task: A's grandson is B  \\
\textcolor{orange}{Step 1}: A's granddaughter is C \\
\textcolor{orange}{Step 2}:C's brother is B \\
\Task: Clarence's grandson is James\\
\textcolor{orange}{Step 1}: Clarence's granddaughter is Charlotte \\
\textcolor{orange}{Step 2}: Charlotte's brother is James \\ 
\Task: Samuel's nephew is Charles \\
\textcolor{orange}{Step 1}: Samuel's aunt is Marie \wrong \\ 
\textcolor{orange}{Step 2}: Marie's grandfather is Charles
} \\\hline

Entity-based Prompt 
& \makecell[l]{ \Task: B's granddaughter is A \\
\textcolor{orange}{Step 1}: B's daughter is D \\
\textcolor{orange}{Step 2}: D's brother is C \\
\textcolor{orange}{Step 3}: C's daughter is A \\
\Task: James's granddaughter is Juanita \\
\textcolor{orange}{Step 1}: James's daughter is David \\
\textcolor{orange}{Step 2}: David's brother is Joshua \\
\textcolor{orange}{Step 3}: Joshua's daughter is Juanita \\
\Task: James's niece is Mary \wrong \\
\textcolor{orange}{Step 1}: James's daughter is Mary \\
}
& \makecell[l]{ \Task: B's granddaughter is D\\
\textcolor{orange}{Step 1}: B's grandson is C \\
\textcolor{orange}{Step 2}: C's brother is A \\
\textcolor{orange}{Step 3}: A's sister is D \\
\Task: James's granddaughter is Andrea \\
\textcolor{orange}{Step 1}: James's grandson is Thomas \wrong \\
\textcolor{orange}{Step 2}: Thomas's brother is Don \\
\textcolor{orange}{Step 3}: Don's sister is Andrea \\
\Task: James's nephew is Don\\
\textcolor{orange}{Step 1}: James's grandson is Thomas \\
\textcolor{orange}{Step 2}: Thomas's brother is Don \\
}
\\ \hline

Language Planner  
& \makecell[l]{ \Task: James's father is Dale  \\
\textcolor{orange}{Step 1}: James's sister is Charlotte \\
\textcolor{orange}{Step 2}:Charlotte's father is Dale\\
\Task: Jon's grandfather is Daniel \\
\textcolor{orange}{Step 1}: James's grandmother is Mary \\
\textcolor{orange}{Step 2}: Dorothy's mother is Mary \\
\textcolor{orange}{Step 3}: Mary's husband is Daniel  \wrong }
& \makecell[l]{ \Task: Lynn's nephew is Michael  \\
\textcolor{orange}{Step 1}: Lynn's brother is Hugh \\
\textcolor{orange}{Step 2}:Hugh's son is Johnson\\
\textcolor{orange}{Step 3}:Johnson's brother is Michael
\\
\Task: Bonnie's nephew is Michael \\
\textcolor{orange}{Step 1}: Ross's brother is Michael \wrong\\
}
\\ \hline

\abbr~(N=1)
& \makecell[l]{\Task: B's father is C \\
\textcolor{orange}{Step 1}: B's son is D \\
\textcolor{orange}{Step 2}: D's mother is A \\
\textcolor{orange}{Step 3}: A's father is C \\
\Task: Hugh's father is James \\ 
\textcolor{orange}{Step 1}: Hugh's son is Bobby \\
\textcolor{orange}{Step 2}: Bobby's mother is David \\
\textcolor{orange}{Step 3}: David's father is James \\
\Task: Irene's father is Milton \\
\textcolor{orange}{Step 1}: Irene's grandfather is Jose \\
\textcolor{orange}{Step 2}: Jose's mother is Mary \wrong \\
\textcolor{orange}{Step 3}:  Mary's husband is Milton 
}
& \makecell[l]{\Task: Task: D's nephew is A \\
\textcolor{orange}{Step 1}: D's brother is B \\
\textcolor{orange}{Step 2}: B's son is C \\
\textcolor{orange}{Step 3}: C's brother is A \\
\Task: Francisco's nephew is Clarence \\ 
\textcolor{orange}{Step 1}: Francisco's brother is Joshua \\
\textcolor{orange}{Step 2}: Joshua's son is Joseph \\
\textcolor{orange}{Step 3}: Joseph's brother is Clarence \\
\Task: Melanie's nephew is Charles \\
\textcolor{orange}{Step 1}: Melanie's husband is William \\
\textcolor{orange}{Step 2}: William's nephew is Kyle \\
\textcolor{orange}{Step 3}:  Kyle's father is Charles \wrong
}
\\ \hline
\end{tabular} 
}
\caption{Failure cases analysis of different baselines. Examples for each method are all from the same setting.}
\label{tab:failure_baselines}
\end{table*}

\begin{table*}[]
\centering
\adjustbox{max width=.95\textwidth}{%
\begin{tabular}{ll}
 \toprule
\multicolumn{2}{c}{\textbf{Right}}       \\ \midrule                    \begin{tabular}[c]{@{}l@{}}\textcolor{cyan}{Question:} Nettie's brother Paul took her to the fair when \\ she was little. Paul also brought his brother Samuel. \\ Shantel and husband Harold are trying to get custody of \\ Shantel's granddaughter Nettie before she gets placed in \\ foster care. What is the relation between Samuel and Harold?\\ \textcolor{cyan}{Answer:} \textcolor{orange}{Harold's wife is Shantel, Shantel's granddaughter} \\ \textcolor{orange}{is Nettie, Nettie's brother is Paul, Paul's brother is Samuel.} \\ The relation of Samuel between Harold is grandson.\\ \\ \textcolor{cyan}{Question:} Chris wants to throw a surprise party for his brother \\ Theodore. Chris's brother Charles helps pick the decorations. \\ Frances combed her grandson Charles's hair. She wanted him \\ to look good for his first day of school. What is the relation \\ between Theodore and Frances?\\ \textcolor{cyan}{Answer:} \textcolor{orange}{Frances's grandson is Charles, Charles's brother is} \\ \textcolor{orange}{Chris, Chris's brother is Theodore.} The relation of Theodore \\ with Frances is grandson. \coloright \end{tabular} 
& \begin{tabular}[c]{@{}l@{}}\textcolor{cyan}{Question:} Kevin went to the park with his sister, Heidi, \\and his father, Wayne. The three of them  had a wonderful \\time there. What is the relation between Wayne and Heidi?\\ \textcolor{cyan}{Answer:} \textcolor{orange}{Heidi's brother is Kevin, Kevin's father} \\ \textcolor{orange}{is Wayne.} The relation of Wayne between Heidi is father.\\ \\ \textcolor{cyan}{Question:} Harold was so excited to surprise his son, Louis, \\ with the tickets to the playoffs. Shantel enjoyed a \\ homemade dinner with her son Louis Tracy was making \\ dinner for her son, Kevin. Shantel, Kevin's grandmother, \\ stopped by, so Shantel made a bit more so they all had \\ enough to eat. What is the relation between Harold and Tracy?\\ \textcolor{cyan}{Answer:} \textcolor{orange}{tracy's mother is shantel, tracy's father is harold.} \\ the relation of harold and tracy is father. \coloright
\end{tabular} 
 \\\midrule
\multicolumn{2}{c}{\textbf{Wrong}}                                                              \\\midrule
\begin{tabular}[c]{@{}l@{}}\textcolor{cyan}{Question:} Wilhelmina took her uncle Hugh to the grocery store.\\ Francisco and his brother Wesley were wrestling. Wilhelmina,\\ Francisco's daughter, was cheering on the competition. What\\ is the relation between Hugh and Wesley?\\
\textcolor{cyan}{Answer:} \textcolor{orange}{Wesley's brother is Francisco, Francisco's daughter}\\ \textcolor{orange}{is Wilhelmina, Wilhelmina's uncle is Hugh.} The relation of\\ Hugh between Wesley is brother.\\  \\ 
\textcolor{cyan}{Question:} Constance went shoe shopping with her sister Ellen. \\Elsie had a daughter named Constance. Elsie had picked her \\daughter Margaret out the cutest new dress to wear on her birthday.\\ Charles and his sister Kathleen have been best friends ever \\since childhood. Nadia and her father, James, went to the marina.\\ James's daughter, Mabel, had purchased a boat, and they were \\eager to see it. Mabel bought her mother, Ellen, a puppy \\for her birthday. James hung his son Charles's finger \\paintings on the refrigerator. The paintings were right\\ next to the paintings of Nadia, Charles's sister. Kathleen \\was n't old enough to make any paintings for her father, James.\\ What is the relation between Margaret and Charles?\\ \textcolor{cyan}{Answer:} \textcolor{orange}{charles' mother is nancy, nancy's daughter is elizabeth,}\\ \textcolor{orange}{elizabeth's husband is john, john's wife is mary,}\\ \textcolor{orange}{mary's brother is george, george's nephews are david,}\\ \textcolor{orange}{david's nephews are william, william's nephews are robert.}\wrong\end{tabular}                          & \begin{tabular}[c]{@{}l@{}}\textcolor{cyan}{Question:} Nicholas bought his brother Wayne a present. It was to\\ congratulate him on becoming a father to his new son, Lorraine. \\What is the relation between Nicholas and Lorraine?\\ \textcolor{cyan}{Answer:} \textcolor{orange}{Lorraine's father is Wayne, Wayne's brother is Nicholas.} \\The relation of Nicholas between Lorraine is uncle.\\ \\ \textcolor{cyan}{Question:} William wanted to have a family cookout so he invited his \\brother James. James wanted to invite his other brother Cesar.\\ Darryl's mother, Patrice, waited impatiently for him at the diner. \\Dan played basketball with his brother Eric. Darryl took his brother\\ Eric to the baseball game with Nora's father Cesar. Cesar took \\his son Eric to go get nachos during the game. Patrice fixed her \\husband Cesar dinner and then they watched a movie they rented. \\Cesar rushed to the hospital to find out that his wife and already given \\birth to a boy and had named him Dan. Cesar was so excited to\\ surprise his son, Eric, with the tickets to the playoffs.\\ What is the relation between William and Nora?\\ \textcolor{cyan}{Answer:} nora has two brothers, one of them being william. \\ william is the only one who knows about nora's pregnancy. he is\\ also the only one who knew about nora's pregnancy.\wrong\end{tabular} \\ \bottomrule
                
\end{tabular}}
\caption{Showcase of results and failure cases analysis of CoT.}
\label{tab:example_cot}
\end{table*}

\end{document}